\documentclass[lettersize,journal]{IEEEtran}
\usepackage{amsmath,amsfonts}
\usepackage{algorithmic}
\usepackage{algorithm}
\usepackage{array}
\usepackage[caption=false,font=normalsize,labelfont=sf,textfont=sf]{subfig}
\usepackage{textcomp}
\usepackage{stfloats}
\usepackage{url}
\usepackage{verbatim}
\usepackage{graphicx}
\usepackage{cite}
\usepackage{multirow, makecell}
\usepackage{bbm}
\usepackage{amsfonts}

\usepackage{hyperref}
\hypersetup{breaklinks=true, colorlinks=true, pdfusetitle=true }

\begin{document}

\title{Boosting Monocular 3D Object Detection with Object-Centric Auxiliary Depth Supervision}

\author{Youngseok Kim, Sanmin Kim, Sangmin Sim, Jun Won Choi, and Dongsuk Kum~\IEEEmembership{Member,~IEEE,}
\thanks{%Manuscript received January 29, 2022; revised August 2, 2022; accepted October 3, 2022.
This work was supported by Institute of Information \& communications Technology Planning \& Evaluation (IITP) grant funded by the Korea government (MSIT) (No.2021-0-00951, Development of Cloud based Autonomous Driving AI learning Software).
\textit{(Corresponding author: Dongsuk Kum.)}} %TODO: 날짜 고치기
\thanks{Youngseok Kim, Sanmin Kim, and Dongsuk Kum are with the Graduate School of Mobility, Korea Advanced Institute of Science and Technology (KAIST), 34141 Daejeon, Republic of Korea, (e-mail: \{youngseok.kim, sanmin.kim, dskum\}@kaist.ac.kr)}
\thanks{Sangmin Sim is with the Automotive R\&D Division, Hyundai Motor Company, 06182 Seoul, Republic of Korea, (e-mail: sm.sim@hyundai.com)}
\thanks{Jun Won Choi is with Department of Electrical Engineering, Hanyang University, 04763 Seoul, Republic of Korea, (e-mail: junwchoi@hanyang.ac.kr)}
% \thanks{Digital Object Identifier 00.0000/TITS.2022.0000000}
}

\markboth{IEEE TRANSACTIONS ON INTELLIGENT TRANSPORTATION SYSTEMS}{Kim \MakeLowercase{\textit{et al.}}: Boosting Monocular 3D Object Detection with Object-Centric Auxiliary Depth Supervision}

\IEEEpubid{0000--0000~\copyright~2022 IEEE}

\maketitle
\begin{abstract}
Recent advances in monocular 3D detection leverage a depth estimation network explicitly as an intermediate stage of the 3D detection network.
Depth map approaches yield more accurate depth to objects than other methods thanks to the depth estimation network trained on a large-scale dataset.
However, depth map approaches can be limited by the accuracy of the depth map, and sequentially using two separated networks for depth estimation and 3D detection significantly increases computation cost and inference time.
In this work, we propose a method to boost the RGB image-based 3D detector by jointly training the detection network with a depth prediction loss analogous to the depth estimation task.
In this way, our 3D detection network can be supervised by more depth supervision from raw LiDAR points, which does not require any human annotation cost, to estimate accurate depth without explicitly predicting the depth map.
Our novel object-centric depth prediction loss focuses on depth around foreground objects, which is important for 3D object detection, to leverage pixel-wise depth supervision in an object-centric manner.
Our depth regression model is further trained to predict the uncertainty of depth to represent the 3D confidence of objects.
To effectively train the 3D detector with raw LiDAR points and to enable end-to-end training, we revisit the regression target of 3D objects and design a network architecture.
Extensive experiments on KITTI and nuScenes benchmarks show that our method can significantly boost the monocular image-based 3D detector to outperform depth map approaches while maintaining the real-time inference speed.

\end{abstract}


%-------------------------------------------------------------------------
\section{Introduction}
\IEEEPARstart{D}{etecting} objects in 3D space is essential in many applications such as augmented reality, robotics, and autonomous driving.
Despite the success achieved in monocular-based 2D object detection \cite{Sun2015, redmon2016you, Lin2017a, tian2019fcos, zhang2020bridging} and LiDAR-based 3D object detection \cite{zhou2018voxelnet, yan2018second, lang2019pointpillars, shi2019pointrcnn, Shi2020a, meng2021towards, yin2021graph}, performance of 3D object detection from a single RGB image \cite{chen2016monocular, ku2019monocular, Brazil2019, Simonelli, Liu, Li2020b, Simonelli2020, ma2021delving} lags considerably behind.
Thanks to the well established CNN-based monocular 2D object detection network, image-based 3D detectors can run in real-time and predict accurate 2D position on 2D image space, however, they suffer from low 3D localization accuracy.
Due to perspective projection, points in 3D space are mapped onto a 2D image plane and the depth information is lost on image pixels, making depth estimation from a monocular image an ill-posed problem.
For this reason, estimating the accurate depth of the object is considered the crux of monocular 3D object detection \cite{ma2021delving, Simonelli2020a}.

\IEEEpubidadjcol
Recent advances in a monocular 3D detector leverage a monocular depth estimation network \cite{godard2017unsupervised, Godard2019, Guizilini2020, Eigen2014, Fu2018, Lee2019} as an intermediate stage of the 3D detector, referred to as depth map approaches \cite{Ding2020a, Ma2020, Simonelli2020a, Weng2019, Ma}.
The typical process of adopting a depth estimation network to monocular 3D object detection framework is as follows.
\textit{First}, the depth estimation network is trained to regress a per-pixel depth map on the depth prediction dataset.
The depth estimation network can be trained in a self-supervised manner \cite{godard2017unsupervised, Godard2019, Guizilini2020} using image reconstruction from sequential images or in a supervised manner \cite{Eigen2014, Fu2018, Lee2019} using a ground truth depth map obtained by range sensors such as LiDAR. 
\textit{Second}, once the depth estimation network is converged on the depth prediction dataset, the depth network predicts the depth map on images of the 3D object detection dataset while network parameters are frozen.
\textit{Finally}, the 3D detector is trained to output a 3D bounding box given pre-computed depth map input, so that the training and inference process is not end-to-end.
The depth map can have representations as image-like depth map \cite{Ma2020}, or RGB-D \cite{manhardt2019roi, Ding2020a}, or pseudo-LiDAR \cite{Wang2019a, Weng2019, Ma}.
When using a depth map, the training depth estimation network does not require human annotation cost, which allows the depth network to be trained on large-scale training data.

However, directly adopting a depth map as an input of the 3D object detection network, as explained above, can lead to the following significant limitations:

\noindent
1) The depth trained by the depth estimation task cannot accurately predict the depth of the foreground object.
In Table \ref{table:depth comparison} and Fig. \ref{fig:depth comparison}, depth errors are compared between the ground truth depth and depth predicted by DORN \cite{Fu2018} on the KITTI object detection \textit{val split}.
The raw points and foreground points denote the depth from all projected LiDAR points and depth on objects.
The depth errors that occurred in the foreground region are massive in almost every metric since the depth estimation aims to minimize the depth error of \textit{every pixel (including background)}, whereas the 3D detector only requires the accurate depth of \textit{foreground objects}.
Therefore, the depth estimation network trained without considering the foreground region can be sub-optimal for 3D object detection.
We emphasize that the depth error on vulnerable road users (\textit{e.g.}, pedestrian and cyclist) is much more significant.

\noindent
2) The depth value estimated from the depth estimation network is deterministic, whereas the estimated depth has uncertainty.
Since the depth estimation task is a challenging and ill-posed problem, the estimated depth on a far distance object is more likely to have a high variance.
Without considering the uncertainty of depth, the clearly recognizable object will have a high confidence score, although the objects are too far to be accurately localized.

\noindent
3) Using two separated networks for depth estimation and 3D object detection is extremely inefficient in terms of both memory and computation.
The 3D object detection network has to be delayed until the depth estimation network outputs a dense depth map, which usually takes around 400$ms$, due to the cascade design.
Furthermore, processing the predicted depth map back to the 3D object detector is unnecessary because the distance to the object is already predicted on the depth map.

%%%%%%%%%%%%%%%%%%%%%%%%%%%%%%%%%%%%%%%%%%%%%%%%%%%%%%%%%%%%%%%%%%%%%%%%%%%%%%%%%%%%%%%%%
\setcounter{table}{0}
\setlength{\tabcolsep}{0.5em}
\begin{table}[!t]
    \caption{The Object-Centric Depth Estimation Performance of DORN \cite{Fu2018}.
    All Denotes Depth Pixels on Three Foreground Classes.}
    \begin{center}
    \resizebox{0.90\columnwidth}{!}
    {
    \begin{tabular}{c|c||c|cccc}
        \hline
         & \multirow{2}{*}{Metric} & \multirowcell{2}{Raw\\Points} & \multicolumn{4}{c}{Foreground Points} \\
         & & & All & Car & Ped. & Cyc. \\
        \hline
        \multirowcell{4}{Lower\\is\\better} & Abs Rel    & 0.109 & 0.180 & 0.170 & 0.220 & 0.250 \\
         & Sq Rel     & 0.677 & 1.221 & 1.125 & 1.603 & 2.445 \\
         & RMSE          & 3.635 & 3.358 & 3.256 & 3.890 & 5.895 \\
         &RMSE$_{log}$& 0.177 & 0.211 & 0.199 & 0.230 & 0.250 \\
        \hline
        \multirowcell{4}{Higher\\is\\better} & $\delta<$1.10 & 0.716 & 0.680 & 0.714 & 0.481 & 0.460 \\
        & $\delta<$1.25 & 0.892 & 0.850 & 0.868 & 0.766 & 0.738 \\
        & $\delta<1.25^2$ & 0.961 & 0.920 & 0.925 & 0.910 & 0.886 \\
        &$\delta<1.25^3$& 0.982 & 0.954 & 0.955 & 0.960 & 0.951 \\
        \hline
    \end{tabular}
    }
    \end{center}
    \label{table:depth comparison}
\end{table}
%%%%%%%%%%%%%%%%%%%%%%%%%%%%%%%%%%%%%%%%%%%%%%%%%%%%%%%%%%%%%%%%%%%%%%%%%%%%%%%%%%%%%%%%%

In this work, we propose a method to boost the RGB image-based 3D object detector, MonoPixel, by leveraging per-pixel depth supervisions in an object-centric manner.
In this way, the 3D detection network can be trained to predict the accurate depth of objects without the limitations of depth map approaches explained above.

We first propose novel \textit{object-centric auxiliary depth loss} to allow the network to learn the lost depth information from RGB images analogous to the depth estimation task.
To train the 3D detection network with a large number of depth supervision, we leverage raw LiDAR points which can be obtained without human annotation cost.
Our object-centric depth loss aims the network to reduce the depth error, particularly on the region where foreground objects exist, to effectively leverage per-pixel depth supervisions for the 3D detection task.
Furthermore, our depth regression model outputs the depth value with its uncertainty to consider the localization uncertainty.

However, since the freely available per-pixel depth supervisions (projected LiDAR points) provide the depth to the \textit{object's surface}, the regression target of the 3D object detector, depth to the \textit{object's center}, cannot be directly supervised.
For effectively training the 3D detector with LiDAR points, we revisit the depth regression target of the 3D detector and reformulate it so that both can have the same regression target.
In this way, our network can directly leverage large-scale per-pixel depth supervision for the 3D object detection task.

The proposed network does \textit{not} explicitly produce a dense depth map during testing time, which enables real-time inference speed.
Unlike the depth map approaches, which compute the depth of every pixel and use the depth map as the input of the 3D detector, our network first detects the center position of object and use the depth value on the corresponding position, which allows our network to run an order of magnitude faster than depth map approaches.

%%%%%%%%%%%%%%%%%%%%%%%%%%%%%%%%%%%%%%%%%%%%%%%%%%%%%%%%%%%%%%%%%%%%%%%%%%%%%%%%%%%%%%%
\setcounter{figure}{0}
\begin{figure}[t]
\begin{center}
\includegraphics[width=0.95\columnwidth]{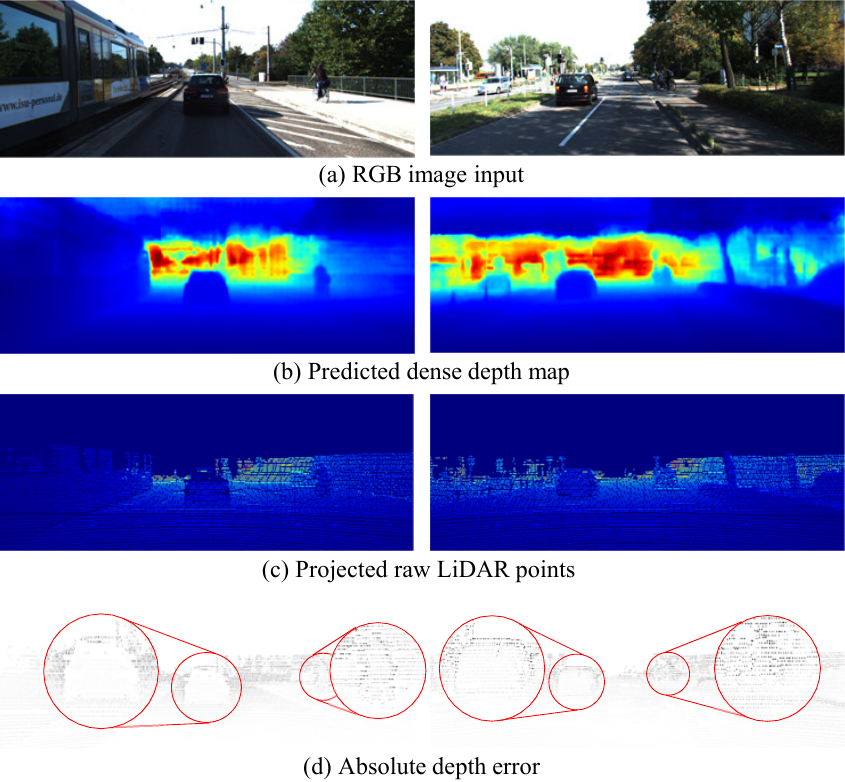}
\end{center}
\caption{Comparison of depth map (b) predicted by depth estimation network DORN \cite{Fu2018} and projected raw LiDAR points (c).
The darker points denote the larger depth error in (d).
The depth error occurred on vulnerable road users are larger than car, although located at a similar distance.
Best viewed in color with zoom in.}
\label{fig:depth comparison}
\end{figure}
%%%%%%%%%%%%%%%%%%%%%%%%%%%%%%%%%%%%%%%%%%%%%%%%%%%%%%%%%%%%%%%%%%%%%%%%%%%%%%%%%%%%%%%

Our extensive experimental results on 3D object detection benchmarks KITTI \cite{Geiger2012} and nuScenes \cite{Caesar2020} show the effectiveness of the proposed boosting method.
In particular, our method achieves a new state-of-the-art on the KITTI dataset compared with existing all RGB image approaches and depth map approaches by a large margin.
On the nuScenes dataset, our method significantly improves the average precision by reducing the translation error, especially on vulnerable road user classes.
Note that our method achieves state-of-the-art performance with a real-time inference speed of 24 FPS.

Our main contributions can be summarized as follows:
\begin{itemize}
  \item We propose a novel object-centric auxiliary depth loss that can significantly boost the 3D detector to learn the accurate depth to the object using raw LiDAR points.
  \item We further introduce object depth decomposition and depth uncertainty guided 3D confidence strategies that can effectively leverage pixel-wise depth supervision for the 3D detection task.
  \item	Extensive experiments on the KITTI and nuScenes object detection benchmark, our method achieves state-of-the-art performance while running in real-time.
\end{itemize}

%-------------------------------------------------------------------------
\section{Related Work}
\subsection{3D object detection from RGB image input}
In advance with 2D object detection, several works \cite{mousavian20173d, chabot2017deep, Brazil2019, liu2019deep, Simonelli, brazil2020kinematic, Simonelli2020} borrow the two-stage anchor-based architecture (\textit{e.g.}, Faster R-CNN \cite{Sun2015}) and directly regress the 3D bounding box from the RGB image input. 
M3D-RPN \cite{Brazil2019} proposes an anchor-based two-stage architecture by utilizing a 3D region proposal.
In \cite{Brazil2019}, 2D-3D anchors are defined by projecting 3D anchors on 2D image space using prior statistics of 3D object size and distance to consider the geometric relationship of the 2D and 3D perspectives.
MonoDIS \cite{Simonelli} proposes a two-stage architecture and trains 3D attributes by disentangled transformation loss to balance multiple regression terms.
Kinematic3D \cite{brazil2020kinematic} expands on M3D-RPN \cite{Brazil2019}, using the baseline to predict kinematic motion and utilizing 3D objects detected within multiple frames with a Kalman filter.
MonoPSR \cite{ku2019monocular} leverages a LiDAR point cloud from an instance to reconstruct the shape and scale information from 2D region proposals.
% revision 1
CaDDN \cite{reading2021categorical} trains the network to predict pixel-wise depth distribution using dense depth map and transforms the front view image features to bird's eye view (BEV) to detect 3D objects on BEV features.
Another group of works \cite{Li2020b, Liu, Chen2020b, ma2021delving} expands the single-stage keypoint-based architecture (\textit{e.g.}, CenterNet \cite{Zhou2019}), using extra keypoints. 
RTM3D \cite{Li2020b} and SMOKE \cite{Liu} expand the center keypoint into 8 vertexes keypoints to represent the 3D bounding box. 
In \cite{Li2020b}, the detected 8 vertexes are further optimized by geometric constraints of the 3D object.
MonoPair \cite{Chen2020b} predicts a center keypoint with an additional pairwise spatial relationship between neighboring objects to post-optimize the regressed distance with a pair-wise distance constraint.

Considering significant achievements have been made in 2D object detectors, \textit{instance-wise supervision} from the 2D bounding box may be sufficient for producing good results for the 2D object detection task.
For example, modern 2D object detectors \cite{Lin2017a, cai2018cascade, tian2019fcos, wang2021scaled} can easily yield over 90\% $AP_{2D}$ on the KITTI object detection dataset \cite{Geiger2012}.
However, the 3D object detection task has much higher complexity, and more supervisions are desired to train the network to accurately predict depth.
We claim that the lack of depth supervision of the RGB image approaches can be the reason for the inferior performance compared to depth map approaches trained by pixel-wise depth map supervisions.

\subsection{3D object detection from depth map input}
Depth map approaches make use of a standalone depth estimation network to make up for the absence of depth information and decouple the challenging depth regression task from the other regression tasks. 
The widely used depth estimation networks for depth map approaches are trained in a supervised manner by the ground truth depth map (\textit{e.g.}, DORN \cite{Fu2018}, BTS \cite{Lee2019}), or in a self-supervised manner by sequential images (\textit{e.g.}, MonoDepth2 \cite{Godard2019}, PackNet-SfM \cite{Guizilini2020}).
Both approaches can benefit from large-scale data (the KITTI depth prediction dataset contains more than 20000 training images) since collecting a ground truth depth map or sequential images does not require human annotation costs.
Several works \cite{Ding2020a, Ma2020} directly take an image-like depth map and processes depth map by a 2D CNN based network in image space.
D4LCN \cite{Ding2020a} processes the depth map by a depth-guided 2D convolution to deal with a scale-variant object size caused by perspective projection.
DDMP \cite{wang2021depth} further integrate the multi-scale depth features with image features on the top of D4LCN architecture via auxiliary depth encoding task.
PatchNet \cite{Ma2020} first predicts the 2D region of interest from the RGB image, crops corresponding patches from the depth map, and predicts 3D location, size, and orientation using a 2D CNN-based network given depth input.
Another line of work \cite{Wang2019a, Weng2019, Ma} transforms the image-like depth map into a point-like pseudo-LiDAR representation to mimic the LiDAR point cloud. 
Pseudo-LiDAR \cite{Wang2019a} adopts an off-the-shelf state-of-the-art LiDAR detector to process the transformed pseudo-LiDAR signal.
MonoPL \cite{Weng2019} and AM3D \cite{Ma} further exploit contextual information from the RGB image using 2D detection or a segmentation sub-network in addition to pseudo-LiDAR.
Some methods \cite{wang2020task, li2020confidence} attempt to improve depth estimation quality for 3D object detection task by focusing more on foreground region.
Specifically, they separately predict depth map of foreground and background region with two sub-networks and combine them to obtain more accurate depth to object.

However, the performance of the depth map approach is bounded to the pre-trained depth estimation network since parameters of the depth estimation network are frozen while training the 3D detection network.
The training objective of the depth estimation task, which is to minimize depth errors of every pixel, including in the background, is sub-optimal to the 3D object detection task that only requires the depth of foreground objects.
Moreover, a recent study \cite{Simonelli2020a} claims that nearly of 30\% images in the KITTI depth prediction \cite{geiger2013vision} \textit{train split} are overlapped with the KITTI 3D object detection \cite{Geiger2012} \textit{val split}.
Since the depth estimation network is trained by the validation images, the depth map approaches have biased results on the KITTI 3D object detection \textit{val split}.

\subsection{3D confidence in monocular 3D object detection}
Recent studies propose probabilistic detection by predicting the uncertainty over the regression target.
MonoPair \cite{Chen2020b} and UR3D \cite{shi2020distance} regress the distance to the object with its heteroscedastic aleatoric uncertainty following Kendall and Gal \cite{kendall2017}.
Predicted distance regression uncertainty is used for post-optimization \cite{Chen2020b} or multiplied by classification uncertainty \cite{shi2020distance} to represent 3D uncertainty.
MonoDIS \cite{Simonelli} and Simonelli \textit{et al.} \cite{Simonelli2020a} explicitly train the confidence of the 3D object in addition to the traditional 2D confidence with self-supervision.
3D confidence is trained by the L1 error between 8 corners of prediction and the ground truth 3D bounding box to output higher confidence when the localization error between two is small. 

Recent attempts to lift the localization error to 3D confidence have only been attempted on two-stage architecture, but its training supervision is limited to the number of 3D bounding box annotations.
Meanwhile, instance-wise uncertainty-aware regression loss \cite{kendall2017} can easily be extended to our approach that trains the depth network by pixel-wise supervision, which allows the network to learn auxiliary depth information by large-scale supervisions.

%-------------------------------------------------------------------------
\section{Preliminaries}

\subsection{Problem Formulation} \label{Problem Formulation}
The goal of the 3D object detection task is to classify and localize 3D objects of interest in 3D space, given a single RGB image and corresponding camera intrinsics. 
The 3D object is parameterized by class $C$, size $w, h, l$, center $x, y, z$, yaw orientation $\psi$, and confidence $p$.
More specifically, we predict the \textit{the 3D center of 3D bounding box} projected on image plane $u, v$, rather than the center of the 2D bounding box.
Given the projected 3D center (called the keypoint) and its depth $d$ on the image coordinate system is transformed into the 3D camera coordinate system as:
\begin{equation}
z=d, x=\frac{(u-{x^{\prime}})\times z}{f_u}, y=\frac{(v-{y^{\prime}})\times z}{f_v},
\label{eq:1}
\end{equation}
where $({x^{\prime}},\ {y^{\prime}})$ and $(f_u, f_v)$ are the principal point and focal length of the camera.
The 3D confidence $p$ indicates the classification-localization coupled probability, which is the \textit{classification} probability of the prediction being an object, and the \textit{localization} probability of the prediction overlapping the ground truth object in the 3D space.

The depth estimation task aims to learn a monocular depth model that predicts the dense depth map $\hat{D}\in{[0, d_{max}]}^{W\times H}$ for every pixel given in an RGB image $I\in{\mathbb{R}}^{W\times H\times 3}$.
The maximum depth estimation range $d_{max}$ is commonly capped up to a fixed range (\textit{e.g.}, 80$m$ in KITTI depth prediction \cite{Eigen2014}).
In a supervised monocular depth estimation setting, the depth estimation network is trained by a ground truth depth map obtained by LiDAR or an RGB-D camera.

%%%%%%%%%%%%%%%%%%%%%%%%%%%%%%%%%%%%%%%%%%%%%%%%%%%%%%%%%%%%%%%%%%%%%%%%%%%%%%%%%%%%%%%
\setcounter{figure}{1}
\begin{figure}[t]
\begin{center}
\includegraphics[width=0.95\columnwidth]{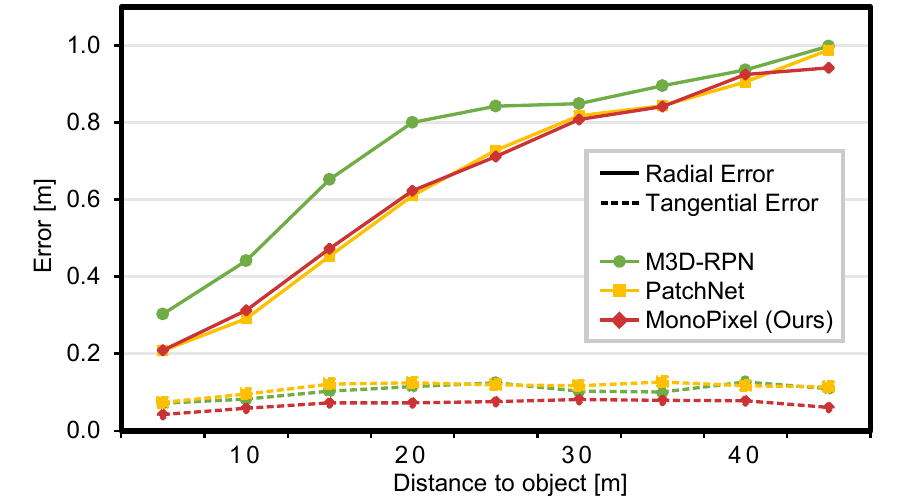}
\end{center}
\caption{Localization error analysis on 2$m$ center-distance matching threshold.}
\label{fig:error analysis}
\end{figure}
%%%%%%%%%%%%%%%%%%%%%%%%%%%%%%%%%%%%%%%%%%%%%%%%%%%%%%%%%%%%%%%%%%%%%%%%%%%%%%%%%%%%%%%

Although both 3D object detection and depth estimation tasks aim to regress depth given in an RGB image, the two tasks have different properties in the region of interest and regression target.
First, 3D detection only requires the depth of the \textit{foreground object}, typically trained by 3D bounding box annotation, whereas the depth estimation requires the depth of \textit{every pixel including background}, trained by per-pixel depth annotation.
Second, the 3D detection task aims to regress the distance to the \textit{center of objects}, but the depth estimation task aims to regress the depth of the \textit{surface of objects}.

\subsection{Key Factor Analysis of Monocular 3D Object Detector}

Recent studies \cite{ma2021delving, Simonelli2020a} provide in-depth analyses that show the key factor of monocular 3D detectors is the localization error using an Average Precision (AP) metric with an IoU (Intersection over Union) matching function.
However, AP and IoU are entangled with multiple metrics (\textit{e.g.}, AP: precision and recall, IoU: localization, size, and orientation), and it is difficult to pinpoint which has the greatest impact.
We further analyze the maximum recall (recall at 0.1 precision) and localization error using the mean absolute error (MAE) in the radial and tangential directions with the center-distance matching function used in nuScenes \cite{Caesar2020}, which is only related to localization.
The radial and tangential directions are defined on the polar coordinate system, which well represents the perspective projection of the pinhole camera model.
In Table \ref{table:performance analysis} and Fig. \ref{fig:error analysis}, we analyze several 3D object detectors that take an RGB image \cite{Brazil2019}, depth map \cite{Ma2020}, and LiDAR points \cite{Shi2020a}.
Experiments were conducted on the KITTI \textit{val split} of \textit{car} \textit{hard} difficulty, and the official code and model provided by the authors.
Note that the number of ground truths within $<$15, 15-30, and 30$<m$ ranges are 2100, 5007, and 3853, respectively. 
%%%%%%%%%%%%%%%%%%%%%%%%%%%%%%%%%%%%%%%%%%%%%%%%%%%%%%%%%%%%%%%%%%%%%%%%%%%%%%%%%%%%%%%
\setcounter{table}{1}
\setlength{\tabcolsep}{0.5em}
\begin{table}[t]
\caption{Performance Analysis of Image and LiDAR 3D Object Detectors}
\centering
\resizebox{1.0\columnwidth}{!}{
    \begin{huge}
    \begin{tabular}{c|c|c||ccc|cc|c}
        \hline
        & & \multirow{3}*{\begin{tabular}{c} Range\\$[m]$ \end{tabular}} & \multicolumn{3}{c|}{Center-distance = $2m$} & \multicolumn{3}{c}{IoU = $0.5$}\\
        \cline{4-9}
        Method & Input & & \multirowcell{2}{Max.\\Recall} & \multirowcell{2}{Rad.\\Err. [$m$]} & \multirowcell{2}{Tang.\\Err. [$m$]} & \multirowcell{2}{Max.\\Recall} & \multirow{2}*{$AP_{3D}$} & \multirow{2}*{$\Delta$} \\
        & & & & & & & &\\ 
         
        \hline
        & \multirow{4}*{RGB} 
         & $<$15 & 94.48 & 0.395 & 0.078 & 75.86 & 68.12 & 7.74\\
        M3D-RPN\footnotemark & & 15-30 & 69.92 & 0.753 & 0.113 & 34.63 & 20.81 & 13.82\\
       \cite{Brazil2019} & & 30$<$ & 46.66 & 0.893 & 0.108 & 21.02 & 6.97 & 14.05\\
        \cline{3-9}
         & & all & 66.20 & 0.692 & 0.102 & 38.49 & 28.59 & 9.90\\
        \hline
        & \multirow{4}*{Depth} 
         & $<$15 & 97.90 & 0.264 & 0.088 & 87.05 & 79.15 & 7.90\\
        PatchNet\footnotemark & & 15-30 & 86.76 & 0.586 & 0.121 & 55.58 & 43.20 & 12.38\\
       \cite{Ma2020} & & 30$<$ & 60.03 & 0.858 & 0.120 & 28.76 & 11.63 & 17.13\\
        \cline{3-9}
         & & all & 79.22 & 0.585 & 0.113 & 53.07 & 39.30 & 13.77\\
        \hline
         & \multirow{4}*{RGB} 
         & $<$15 & 98.57 & 0.279 & 0.053 & 90.90 & 89.57 & 1.33\\
        MonoPixel & & 15-30 & 90.73 & 0.593 & 0.073 & 61.79 & 51.70 & 10.09\\
        (Ours) & & 30$<$ & 65.35 & 0.853 & 0.079 & 35.92 & 16.95 & 18.97\\
        \cline{3-9}
         & & all & 83.31 & 0.596 & 0.072 & 58.28 & 51.50 & 6.78\\
        \hline
        & \multirow{4}*{LiDAR} 
         & $<$15 & 99.33 & 0.078 & 0.040 & 98.90 & 99.25 & 0.35\\
        PV-RCNN\footnotemark & & 15-30 & 98.80 & 0.098 & 0.064 & 97.66 & 97.76 & 0.10\\
       \cite{Shi2020a} & & 30$<$ & 89.26 & 0.145 & 0.092 & 91.75 & 82.43 & 9.32\\
        \cline{3-9}
         & & all & 95.49 & 0.110 & 0.069 & 96.30 & 94.33 & 1.97\\
        \hline
    \end{tabular}
    \end{huge}
    }
\label{table:performance analysis}
\end{table}

\renewcommand{\thefootnote}{1}
\footnotetext{: \url{https://github.com/garrickbrazil/M3D-RPN}}
\renewcommand{\thefootnote}{2}
\footnotetext{: \url{https://github.com/xinzhuma/patchnet}}
\renewcommand{\thefootnote}{3}
\footnotetext{: \url{https://github.com/open-mmlab/OpenPCDet}}
%%%%%%%%%%%%%%%%%%%%%%%%%%%%%%%%%%%%%%%%%%%%%%%%%%%%%%%%%%%%%%%%%%%%%%%%%%%%%%%%%%%%%%%

Our observations are twofold:

1) We found that the major source of the localization error of the monocular 3D detector is dominated by the radial error.
The radial error that occurred on the RGB image approach \cite{Brazil2019} is 18\% larger compared to the depth map approach \cite{Ma2020}, which benefits from the depth estimation network \cite{Fu2018}.
Intuitively, a more accurate depth estimation on objects leads to a higher recall of the 3D detector.
Our method produces a depth as accurate as depth map approaches without using the depth estimation network.

2) The monocular 3D detector suffers from low precision, suggested by the large gap ($\Delta$) between the maximum recall and AP.
Considering the 3D confidence score is the coupled probability of classification and localization, the prediction with higher confidence should have a higher classification probability and lower localization error.
In other words, the 3D score should be high only when the prediction can match the positive matching criteria (\textit{e.g.}, 0.5 IoU).
However, a clearly visible object on the image often has a high confidence score even though the object has a large distance error and is considered as false positive, which harms the precision.
Our proposed method alleviates the problem by improving precision, especially in the near 15$m$ range.
Our method has a higher gap in the far range since it has a much higher maximum recall.

%-------------------------------------------------------------------------
\section{Methodology}

\subsection{Architecture} \label{Architecture}
The proposed architecture is intentionally designed to be simple for autonomous driving, which requires small memory and fast inference speed.
Our framework is extended from a single-stage anchor-free detector CenterNet \cite{Zhou2019} with DLA-34 \cite{Yu2018} and a deformable convolution \cite{dai2017deformable} backbone, which has a good speed-accuracy trade-off.
The backbone network takes an RGB image input $I\in{\mathbb{R}}^{W\times H\times 3}$ and outputs a four times down-sampled feature map $F\in{\mathbb{R}}^{w\times h\times f}$, where $w,h$ are $\frac{W}{4}$, $\frac{H}{4}$ and $f$ is the channel of the feature map. 
On the top of the shared feature map $F$, convolutional sub-networks regress seven 3D attributes and two 2D attributes of the 3D bounding box, as illustrated in Fig. \ref{fig:overall architecture}. 
All convolutional sub-networks have 3×3 and 1×1 convolutional layers with ReLU non-linearity.
We refer the reader to \cite{Zhou2019} for more details of the baseline network.

%%%%%%%%%%%%%%%%%%%%%%%%%%%%%%%%%%%%%%%%%%%%%%%%%%%%%%%%%%%%%%%%%%%%%%%%%%%%%%%%%%%%%%%
\setcounter{figure}{2}
\begin{figure*}[t]
\begin{center}
\includegraphics[width=0.95\textwidth]{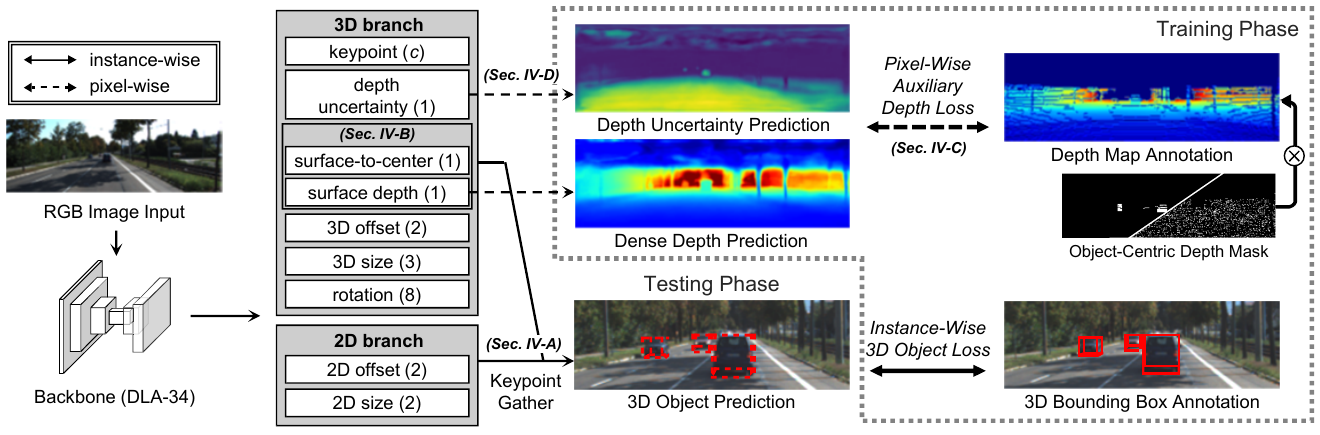} 
\end{center}
\caption{Overview of our framework. 
Our network predicts seven 3D attributes and two 2D attributes in parallel and train them by instance-wise (solid line) loss at the keypoint location.
The surface depth and depth uncertainty sub-networks are additionally trained by pixel-wise (dashed line) loss as well as the instance-wise loss during training.}
\label{fig:overall architecture}
\end{figure*}
%%%%%%%%%%%%%%%%%%%%%%%%%%%%%%%%%%%%%%%%%%%%%%%%%%%%%%%%%%%%%%%%%%%%%%%%%%%%%%%%%%%%%%%

The main contribution of our paper comes from the novel training loss.
During the training phase, the ground truth keypoint heatmap $Y\in[0,1]^{w\times h\times C}$ with Gaussian kernels is produced at the \textit{projection of 3D centers} for each class.
The Gaussian radius is set to $\sigma=\max(f(wl), 0)$ where $f$ is a radius function defined in CornerNet \cite{law2018cornernet} and $w, l$ are the size of the 2D bounding box.
The training objective of the keypoint sub-network is a penalty-reduced pixel-wise logistic regression with focal loss \cite{Lin2017a}:
\begin{equation}
\mathcal{L}_{k}=
\frac{-1}{N}\sum_{x,y,c=1}^{w,h,C}
    \begin{cases}
        (1-\hat{Y}_{xyc})^\alpha \log(\hat{Y}_{xyc})  &\text{if } Y_{xyc}=1\\
        \hfil (1-Y_{xyc})^\beta (\hat{Y}_{xyc})^\alpha \\ 
        \hfil \log(\hat{Y}_{xyc}) &\text{otherwise,}
    \end{cases}
\label{eq:keypoint loss}
\end{equation}
where $\alpha$ and $\beta$ are hyper-parameters of the focal loss, and $N$ is the number of instances in the image.
General modern object detectors train instance-wise attributes (\textit{i.e.}, depth, 3D offset, 3D size, rotation) only at the keypoint since the only available supervision is the 3D bounding box annotation (solid lines in Fig. \ref{fig:overall architecture}).
However, our method additionally trains the surface depth and depth uncertainty sub-networks by pixel-wise auxiliary depth supervisions (dashed lines in Fig. \ref{fig:overall architecture}) using raw LiDAR points.
Details of the proposed pixel-wise auxiliary depth loss will be described in Sec. \ref{Pixel-Wise Depth Supervisions}.

At the testing phase, the network outputs the keypoint heatmap $\hat{Y}\in[0,1]^{w\times h\times C}$, where $C$ is the number of class and regression maps of 3D attributes (\textit{e.g.}, depth $\hat{D}\in[0,d_{max}]^{w\times h}$, 3D size $\hat{S}_{3D}\in\mathbb{R}^{w\times h\times 3}$) in parallel.
Given the predicted keypoint heatmap $\hat{Y}$, the local maximum pixel which corresponds to the center of the 3D object is first predicted, and its value $\hat{y}$ corresponds to a 2D \textit{classification} confidence.
Once keypoint $\hat{y}$ is detected from heatmap $\hat{Y}$, all 3D attributes are gathered from the regression maps at the keypoint location and finally the 3D bounding box is retrieved from the gathered 3D attributes.

%%%%%%%%%%%%%%%%%%%%%%%%%%%%%%%%%%%%%%%%%%%%%%%%%%%%%%%%%%%%%%%%%%%%%%%%%%%%%%%%%%%%%%%%%
\setcounter{table}{2}
\setlength{\tabcolsep}{0.4em}
\begin{table}[!t]
    \caption{IoU Caused by Surface-to-Center Error on KITTI Dataset}
    \begin{center}
    \resizebox{0.95\columnwidth}{!}{
    \begin{tabular}{cc||cccccccc}
        \hline
        \multirow{2}*{Class} &\multirowcell{2}{Radial\\Err. [$m$]} & \multirow{2}*{0.10} & \multirow{2}*{0.15} & \multirow{2}*{0.20} & \multirow{2}*{0.27} & \multirow{2}*{0.59} & \multirow{2}*{0.70} & \multirow{2}*{1.50} & \multirow{2}*{2.00}\\
         & & & & & & & & & \\
        \hline
        Car & \multirowcell{3}{IoU [\%]} & 0.95 & 0.93 & 0.90 & 0.87 & 0.74 & \textbf{0.70} & 0.44 & 0.32\\
        Ped. & & 0.78 & 0.68 & 0.60 & \textbf{0.50} & 0.15 & 0.07 & 0 & 0\\
        Cyc. & & 0.89 & 0.84 & 0.80 & 0.73 & \textbf{0.50} & 0.43 & 0.08 & 0\\
        \hline
    \end{tabular}}
    \end{center}
    \label{table: IoU error}
\end{table}
%%%%%%%%%%%%%%%%%%%%%%%%%%%%%%%%%%%%%%%%%%%%%%%%%%%%%%%%%%%%%%%%%%%%%%%%%%%%%%%%%%%%%%%%%

%-------------------------------------------------------------------------
\subsection{Revisiting Depth Regression Target of 3D Object}
Here, we revisit the regression target of the 3D object detection and depth estimation task.
In general, monocular 3D object detectors \cite{Zhou2019, Brazil2019, Simonelli} predict the \textit{depth to the object's center} $d_c$ with the 3D bounding box annotation. 
The 3D bounding box annotation for autonomous driving application is generally labeled on LiDAR points, which captures the \textit{depth to the object's surface} $d_s$.
Due to the dimension of the object, the distance between the object's surface $d_s$ and center $d_c$, referred to as surface-to-center distance $d_{s2c}$, can be nearly up to half of the object's length, as illustrated in Fig. \ref{fig:object depth decomposition}.
If the network predicts a 3D bounding box with the exact size and orientation as the ground truth but has the depth error as $d_{s2c}$ in a radial direction, the IoU will be 0.33.
As shown in Table \ref{table: IoU error}, the IoU error caused by the radial distance error is non-trivial, considering the matching criteria of the KITTI 3D object detection dataset.
In KITTI, IoU thresholds for Car, Pedestrian, and Cyclist classes are 0.7, 0.5, and 0.5, respectively; thus, the translation error of 0.7$m$, 0.27$m$, and 0.59$m$ in the radial direction can lead to a false positive.
We use an average length of each class on KITTI as 3.9$m$, 0.8$m$, and 1.76$m$ to calculate IoU error.

%%%%%%%%%%%%%%%%%%%%%%%%%%%%%%%%%%%%%%%%%%%%%%%%%%%%%%%%%%%%%%%%%%%%%%%%%%%%%%%%%%%%%%%%%
\setcounter{figure}{3}
\begin{figure}[t]
\begin{center}
\includegraphics[width=0.8\columnwidth]{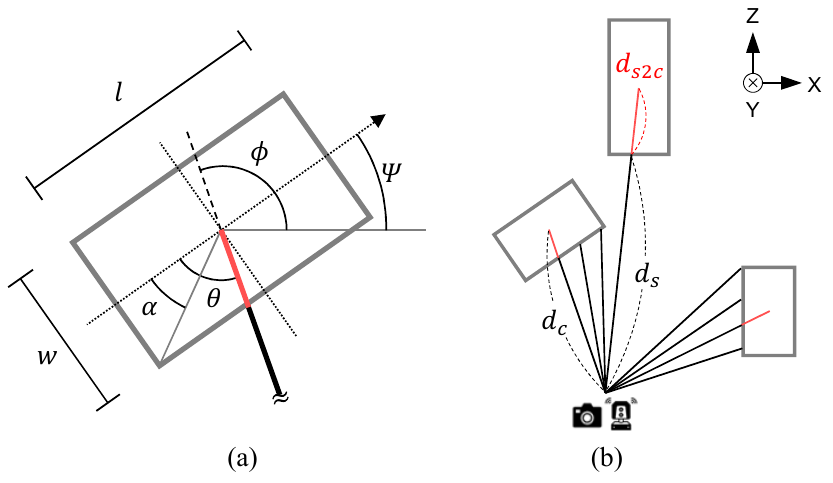}
\end{center}
\caption{Illustration of object depth decomposition. 
The surface-to-center distance of object can be derived from object's 3D bounding box as (a).
Consequently, the distance to object center $d_c$ can be decomposed into surface depth $d_s$ and surface-to-center distance $d_{s2c}$ as (b).
The black lines in (b) denote LiDAR beams in top-view, corresponds to depth pixels of the depth map in image-view.}
\label{fig:object depth decomposition}
\end{figure}
%%%%%%%%%%%%%%%%%%%%%%%%%%%%%%%%%%%%%%%%%%%%%%%%%%%%%%%%%%%%%%%%%%%%%%%%%%%%%%%%%%%%%%%%%

To remedy the inconsistency between two regression targets, the depth of the \textit{3D object's center} $d_c$ and depth of the \textit{object's surface} $d_s$, we propose the \textit{object depth decomposition} strategy.
\begin{equation}
d_c=d_s+d_{s2c}
\label{eq:3}
\end{equation}
By defining the depth to the 3D center of object $d_c$ by surface depth $d_s$ and surface-to-center distance $d_{s2c}$, the regression target of the 3D object detector becomes consistent with the per-pixel depth supervisions.
More specifically, the surface-to-center distance $d_{s2c}$ can be derived from the 3D size $w,l$, direction $\phi$, and yaw orientation $\psi$ of object as:
\begin{equation}
d_{s2c}=
    \begin{cases}
        (l/2)/cos(\theta)  &\text{if } \theta \leq \alpha\\
        (w/2)/cos(\pi/2-\theta) &\text{otherwise,}
    \end{cases}
\label{eq:4}
\end{equation}
where $\theta=\lvert \psi-\phi \lvert$, $\phi=arctan(z/x)$, and $\alpha = arctan(w/l)$.
The direction to the object $\phi$ can be further derived as a function of the pixel $u$ as $\phi=arctan(z/x)=arctan(f_u/(u-x^{\prime}))$ using Eq. \ref{eq:1}.
This indicates that the surface-to-center distance $d_{s2c}$ does not depend on the depth of the object but is determined by its size and orientation.

Considering the surface depth can vary from 0 to 60$m$, we predict the inverse depth and transform it into the absolute depth by inverse sigmoid transformation $d_s = 1/\sigma(\hat{d_s})-1$.
In contrast, the surface-to-center distance $d_{s2c}$ is directly regressed as a real value in meters.
The surface-to-center distance $d_{s2c}$ is only assigned at the object keypoint as shown by the red line in Fig. \ref{fig:object depth decomposition}(b), while more than one surface depth $d_s$ can be trained on the object, shown as black lines.
Note that surface depth $d_s$ at the object keypoint is always calculated using the ground truth as $d_c-d_{s2c}$, where the center depth $d_c$ and surface-to-center distance $d_{s2c}$ are derived from the 3D annotation.
This can handle the situation that the surface depth from LiDAR points can be inconsistent to actual distance to object's surface when object's center is occluded.

%%%%%%%%%%%%%%%%%%%%%%%%%%%%%%%%%%%%%%%%%%%%%%%%%%%%%%%%%%%%%%%%%%%%%%%%%%%%%%%%%%%%%%%%%
\setcounter{figure}{4}
\begin{figure}[t]
\begin{center}
\includegraphics[width=1.0\columnwidth]{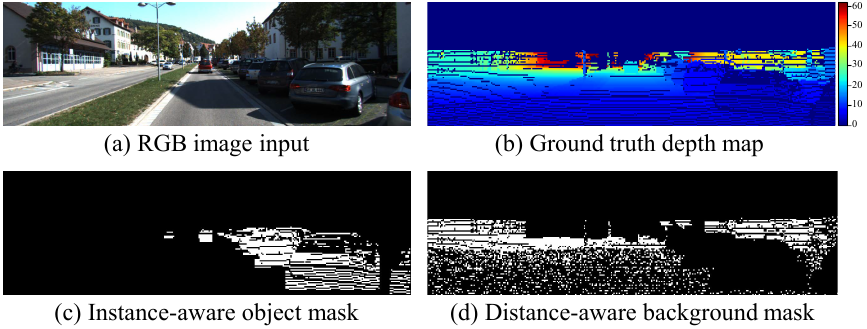}
\end{center}
\caption{Examples of depth map annotation and object-centric depth masks.
Note that the depth map and depth masks are four times down-sampled from original image size to have a same size as a feature map.}
\label{fig:depth mask}
\end{figure}
%%%%%%%%%%%%%%%%%%%%%%%%%%%%%%%%%%%%%%%%%%%%%%%%%%%%%%%%%%%%%%%%%%%%%%%%%%%%%%%%%%%%%%%%%

%-------------------------------------------------------------------------
\subsection{Pixel-Wise Depth Supervisions for 3D detector} \label{Pixel-Wise Depth Supervisions}
We decompose the regression target of the 3D object to have a surface depth $d_s$ to leverage the depth value of LiDAR points to train the 3D detection network.
By doing this, the number of depth supervisions for training the 3D detector can be significantly improved to the number of LiDAR points from the number of 3D objects in the image.
However, depths on foreground objects occupy only a small area in the image, \textit{i.e.}, depth pixels that lie on the car class on KITTI are 11.42\% of the total depth pixels, which can be easily diminished by the overwhelming background pixels. 
To this end, we design an object-centric depth loss to focus on training the depth of foreground objects while still benefiting from large-scale pixel-wise depth supervisions.
Note that per-pixel depth supervisions are only used for training and not used during the inference phase.

\subsubsection{Object-Centric Depth Masking} \label{Object-Centric Depth Masking}
To alleviate the overwhelming depth pixels from background pixels, the network should be trained to focus more on the important supervisions for the 3D object detection.
The depth pixels on foreground objects are first masked as an object mask, as shown in Fig. \ref{fig:depth mask}(c).
In our experiment, we mask LiDAR points inside the 3D bounding box annotation as the foreground depth pixel, then project them into the image plane to obtain depth map and corresponding depth mask.
We expect that the 2D bounding box, which is less expensive to annotate, can also be used for object masking.
Additionally, the number of LiDAR points per area decreases inversely proportional to the square of the distance due to the nature of the LiDAR mechanism.
We divided the background depth at 10$m$ range intervals and sub-sample the depth pixels by distance to normalize the depth value distribution as shown in Fig. \ref{fig:depth mask}(d).
We denote the indicator functions for foreground and background masking by $\mathbbm{1}_{fg}$ and $\mathbbm{1}_{bg}$, being 1 if the pixel is masked and 0 otherwise.

\subsubsection{Object-Centric Auxiliary Depth Loss}
Our depth-related sub-networks (\textit{i.e.}, surface depth and depth uncertainty) are jointly trained by \textit{instance-wise} 3D bounding box supervision and \textit{pixel-wise} depth map supervision with two loss terms to leverage the depth map in an object-centric manner.
The uncertainty-aware regression loss \cite{kendall2017} is adopted to regress the depth $\hat{d}$ with the heteroscedastic aleatoric uncertainty as variance $\hat{\sigma}^2$.

The loss for training depth-related sub-networks can be formulated as follows. 
The regression model $f^\theta$ takes an input image $I\in{\mathbb{R}}^{W\times H\times3}$ and predicts $w\times h$ dense depth $\hat{D}\in[0,80]^{w\times h}$ and its variance $\hat{\Sigma}^2\in{\mathbb{R}}^{w\times h}$ as:
\begin{equation}
[\hat{D}, \hat{\Sigma}^2]=f^\theta(I).
\label{eq:6}
\end{equation}
Since the ground truth keypoint location is known during training, the depth $\hat{d}\in[0,80]^N$ and variance  $\hat{\sigma}^2\in{\mathbb{R}}^N$ at \textit{N object keypoint location} can be extracted using a gathering operation.
The \textit{instance-wise} depth loss is formulated as:
\begin{equation}
\mathcal{L}_{dep}^{obj}=\frac{1}{N}\sum_{i=1}^{N}\left[\frac{\lVert d_{i}^*-\hat{d}_{i}\rVert_1}{\hat{\sigma}_i^2} + \log \hat{\sigma}_i^2\right],
\label{eq:instance-wise depth loss}
\end{equation}
where $N$ denotes the number of ground truth 3D bounding boxes on the image and $d_{i}^*$ is the ground truth surface depth at the object keypoint location.

The uncertainty-aware regression loss can be applied to \textit{pixel-wise} dense regression without loss of generality as:
\begin{equation}
\mathcal{L}_{dep}^{m}=\frac{1}{N_{m}}\sum_{x,y=1}^{w,h}{\mathbbm{1}_m\left[\frac{\lVert{D_{(x,y)}^*-\hat{D}_{(x,y)}}\rVert_1}{\hat{\Sigma}^2_{(x,y)}} + \log(\hat{\Sigma}^2_{(x,y)})\right]},
\label{eq:pixel-wise depth loss}
\end{equation}
where $N_{m}$ denotes the number of masked \textit{pixels}, $D_{(x,y)}^*$ is the ground truth depth map, and $\mathbbm{1}_m$ is the indicator function.
The depth values on the depth map $D_{(x,y)}^*$ do not always lie exactly on the center of the object $d_{i}^*$, but it provides strong auxiliary supervision.
In practice, the variance $\hat{\sigma}^2$ is predicted as log variance $s:=\log\hat{\sigma}^2$ for the numerical stability.

By using indication functions for the foreground $\mathbbm{1}_{fg}$ and background $\mathbbm{1}_{bg}$ defined in Sec. \ref{Object-Centric Depth Masking}, our depth regression model is supervised with \textit{instance-wise} loss and object-centrically weighted auxiliary \textit{pixel-wise} loss:
\begin{equation}
\mathcal{L}_{dep}^{tot}=\mathcal{L}_{dep}^{obj} + \lambda \mathcal{L}_{dep}^{fg} + (1 - \lambda) \mathcal{L}_{dep}^{bg},
\label{eq:9}
\end{equation}
where $\lambda$ is the foreground mask weighting parameter.

%-------------------------------------------------------------------------
\subsection{Depth Uncertainty Guided 3D Object Confidence}
The problem of confidence defined in 2D image space as Eq. \ref{eq:keypoint loss} is that confidence only implies the certainty of 2D localization in the image space, while the confidence in the depth direction cannot be considered.
For example, a clearly visible but distant object is prone to have a high 2D confidence score, even if the depth to the object has high uncertainty.
As formulated in Eq. \ref{eq:1}, the 3D position of the object ($x, y, z$) can be derived from the 2D pixel ($u, v$) and depth $d$, if camera intrinsic parameters are given. 
Therefore, the localization probability of the 3D object can be explicitly decomposed into the probability of prediction being a class of interest, probability of prediction well localized in 2D image space, and depth direction.
In our 3D detection problem setting, detecting the keypoint of objects represents the combination of the classification and 2D localization probability.
Best of all, the depth confidence is already provided in an uncertainty form, which also makes the most of pixel-wise depth supervision by Eq. \ref{eq:pixel-wise depth loss}.

The depth variance $\sigma^2\in\mathbb{R}^N$ of an object estimated by Eq. \ref{eq:6} can be mapped into the confidence form $p_{dep}\in[0,1]^N$ via the negative exponential function:
\begin{equation}
p_{dep}=e^{-\sigma^2},
\label{eq:10}
\end{equation}
which produces lower confidence when the estimated depth has higher variance.
The keypoint confidence $p_{k}$ is trained by Eq. \ref{eq:keypoint loss} and depth confidence $p_{dep}$ is trained by Eq. \ref{eq:9}.
To further represent the localization probability in the depth direction, our 3D confidence can be represented as a combination of the keypoint and depth confidences by the law of total probability as:
\begin{equation}
p_{3D}=p_{k}p_{dep}.
\label{eq:3d confidence}
\end{equation}

%-------------------------------------------------------------------------
\section{Experiments}

\subsection{Implementation Details} \label{Implementation Details}
\subsubsection{Dataset and Metrics}
We evaluate our method on KITTI \cite{Geiger2012} and nuScenes \cite{Caesar2020} 3D object detection benchmarks. 
The KITTI dataset consists of 7481 training and 7518 testing images with 3D annotation of Car, Pedestrian, and Cyclist classes.
We split 7481 training images into 3712 \textit{train split} and 3769 \textit{val split}, following Chen \textit{et al.} \cite{Chen2015} for training and ablation experiments.
An average precision with 40 recall points ($AP_{40}$ \cite{Simonelli}) with 3D and bird's eye view Intersection-over-Union (IoU) matching criteria is used as an evaluation metric.
The default IoU thresholds for each class are set to 0.7, 0.5, and 0.5.
KITTI defines the difficulty of each ground truth as easy, moderate, and hard based on the size, occlusion, and truncation levels.
The depth estimation performance was further evaluated using the metrics described in Eigen \textit{et al.} \cite{Eigen2014}, and the mean absolute error (MAE) with IoU and center-distance matching for comprehensive evaluation.

nuScenes consists of 1000 sequences, with 700, 150, and 150 sequences for \textit{training}, \textit{validation}, and \textit{testing}, respectively.
Each sequence has approximately 40 frames with a 2Hz framerate, and each frame contains synchronized 6 camera images and 360 degree LiDAR points with 3D bounding box annotation.
The annotations include 10 classes and are evaluated by $AP$ with a center-distance matching at 0.5$m$, 1$m$, 2$m$, and 4$m$ instead of the IoU.
Additionally, we report true positive metrics, average translation error (ATE), average scale error (ASE), and average orientation error (AOE) following nuScenes \cite{Caesar2020}.
Note that $TP$ metrics are calculated using a 2$m$ center distance threshold.

\subsubsection{Preprocessing}  
We preprocess the ground truth depth map for training by projecting LiDAR points into the image using camera intrinsic extrinsic. 
The input image and corresponding depth map are resized into $1280\times384$ for KITTI and $800\times448$ for nuScenes.
Since the network outputs four times down-sampled size feature map given image input, we down-sample the ground truth depth map to have the same size as the feature map.
To this end, we use $4\times4$ min-pooling to preserve the closer depth pixels that are more likely to be the foreground object.

\subsubsection{Data Augmentation}  
We apply random horizontal flipping, scaling, cropping, and color jittering for both the image and depth map during the training phase. 
The depth value of the depth map and 3D bounding box is also scaled corresponding to the image scaling ratio.
Horizontal flipping is used for the test time augmentation (TTA) following \cite{Zhou2019}.

\subsubsection{Training details}
Depth map approaches adopt the depth estimation network trained on the KITTI depth prediction dataset with Eigen split \cite{Eigen2014}, which consists of 23158 training images.
However, as claimed by \cite{Simonelli2020a}, nearly a third of the KITTI object detection \textit{val split} images are from the KITTI depth prediction \textit{train split}.
Moreover, the depth prediction dataset does not prioritize collecting dynamic objects (\textit{e.g.}, car, pedestrian), and many frames only contain monotonous backgrounds in terms of the 3D object detection task .
To avoid the data overlapping issue and leverage the diverse depth data, we decided to utilize the depth map from the nuScenes \textit{train set}.
We only use front camera images among the 6 cameras, which consists of 28130 training images, similar to the number of KITTI depth prediction images.
Note that nuScenes uses a 32 beams LiDAR, which produces a more sparse depth map compared to the 64 beams LiDAR of KITTI.

%%%%%%%%%%%%%%%%%%%%%%%%%%%%%%%%%%%%%%%%%%%%%%%%%%%%%%%%%%%%%%%%%%%%%%%%%%%%%%%%%%%%%%%%%
\setlength{\tabcolsep}{0.6em} % for the horizontal padding
\setcounter{table}{3}
\begin{table*}[!t]
    \caption{$AP_{40}$ Performance on KITTI Test Set for Car Class at 0.7 IoU. 
    The Best and Second Best Results Are in \textbf{bold} and \underline{underline}. 
    Input Data Denotes the Type of Data or External Network for Training and Inference and Extra Data Denotes the Additional Data used for Training.
    Seg, 2D Det, Seq, and LiDAR Denote Instance Segmentation, 2D Detector, Sequential RGB Images, and LiDAR Points.
    $^\dagger$ Denotes the Runtime Reported on KITTI Leaderboard.}
    \begin{center}
    \resizebox{0.8\textwidth}{!}{
    \begin{tabular}{c|c|c||ccc|ccc|c}
        \hline
        \multirow{2}*{Method} & \multirow{2}*{Input Data} & Extra & \multicolumn{3}{c|}{Car, $AP_{3D}$ [\%]} & \multicolumn{3}{c|}{Car, $AP_{BEV}$ [\%]} & {Time} \\
        & & Data & Easy & Mod. & Hard & Easy & Mod. & Hard & [ms]\\
        \hline
        MonoPL \cite{Weng2019}  & Depth, Seg& - & 10.76 & 7.50 & 6.10 & 21.27 & 13.92 & 11.25 & - \\
        AM3D \cite{Ma}          & Depth, RGB, 2D Det& - & 16.50 & 10.74 &  9.52 & 25.03 & 17.32 & 14.91 & 400$^\dagger$ \\
        PatchNet \cite{Ma2020}  & Depth, 2D Det& - & 15.68 & 11.12 & 10.17 & 22.97 & 16.86 & 14.97 & 486 \\
        D4LCN \cite{Ding2020a}  & Depth, RGB& - & 16.65 & 11.72 & 9.51 & 22.51 & 16.02 & 12.55 & 200$^\dagger$ \\
        Demystifying \cite{Simonelli2020a} & Depth, 2D Det& - & 22.40 & 12.53 & 10.64 & - & - & - & 486 \\
        Kinematic3D \cite{brazil2020kinematic} & Seq & - & 19.07 & 12.72 & 9.17 & 26.69 & 17.52 & 13.10 & 120 \\
        DDMP\cite{wang2021depth} & Depth, RGB & Depth & 19.71 & 12.78 & 9.80 & 28.08 & 17.89 & 13.44 & 180$^\dagger$ \\
        \hline
        MonoPSR \cite{ku2019monocular}  & RGB, 2D Det & LiDAR & 10.76 & 7.25 & 5.85 & 18.33 & 12.58 & 9.91 & 200 \\
        MonoDIS \cite{Simonelli}        & RGB & - & 10.37 & 7.94 & 6.40 & 17.23 & 13.19 & 11.12 & - \\
        M3D-RPN \cite{Brazil2019}       & RGB & - & 14.76 & 9.71 & 7.42 & 21.02 & 13.67 & 10.23 & 161 \\
        SMOKE \cite{Liu}                & RGB & - & 14.03 & 9.76 & 7.84 & 20.83 & 14.49 & 12.75 & \textbf{30} \\
        MonoPair \cite{Chen2020b}       & RGB & - & 13.04 & 9.99 & 8.65 & 19.28 & 14.83 & 12.89 & 57 \\
        RTM3D \cite{Li2020b}            & RGB & - & 14.41 & 10.34 & 8.77 & 19.17 & 14.20 & 11.99 & 55 \\
        MoVi-3D \cite{Simonelli2020}    & RGB & - & 15.19 & 10.90 & 9.26 & 22.76 & 17.03 & 14.85 & 45 \\
        MonoDLE \cite{ma2021delving}    & RGB & - & 17.23 & 12.26 & 10.29 & 24.79 & 18.89 & 16.00 & \underline{40} \\
        
        CaDDN \cite{reading2021categorical} & RGB & Depth & \textbf{19.17} & \textbf{13.41} & 11.46 & 27.94 & 18.91 & 17.19 & 630$^\dagger$ \\
        \hline
        MonoPixel (Ours) & RGB & LiDAR & \underline{19.06} & \underline{13.33} & \textbf{11.90} & \textbf{27.98} & \textbf{19.75} & \textbf{17.32} & 42 \\
        % Improvement & - & +1.83 & +1.07 & +1.61 & +3.19 & +0.86 & +1.32 & - \\
        \hline
    \end{tabular}}
    \end{center}
    \label{table:kitti car}
\end{table*}
%%%%%%%%%%%%%%%%%%%%%%%%%%%%%%%%%%%%%%%%%%%%%%%%%%%%%%%%%%%%%%%%%%%%%%%%%%%%%%%%%%%%%%%%%

For KITTI pre-training, we only train the surface depth and depth uncertainty sub-network with auxiliary \textit{pixel-wise} depth loss, as in Eq. \ref{eq:pixel-wise depth loss}, which does not require \textit{instance-wise} supervision, \textit{i.e.}, without a 3D bounding box.
After the model is pre-trained for 70 epochs on nuScenes depth, it is then trained another 70 epochs on KITTI with full loss.
For nuScenes, the model is trained for 140 epochs in an end-to-end manner.
We use the AdamW \cite{loshchilov2018decoupled} optimizer, a batch size of 32 on 4 GPUs, and a learning rate of $1.25e-4$ decayed by 0.1 at the 90th and 120th epoch on nuScenes, and the 45th and 60th epoch on KITTI. 
The inference time was measured on an Intel i9-10980XE CPU and NVIDIA RTX 3090 GPU.

\subsubsection{Hyperparameters}
The loss weights for regression sub-networks are set to 1, except for the 2D size sub-network set to 0.1.
We use $\alpha$=2 and $\beta$=4 for keypoint focal loss, and the foreground mask weight $\lambda$ is set to 0.7 for object-centric auxiliary depth loss.
Invalid predictions are filtered out with a 2D confidence score lower than 0.4 and 0.1 before multiplying depth confidence during testing for KITTI and nuScenes.
For nuScenes, 3D detection is performed separately on a single image and detection results from six camera images are transformed into the vehicle coordinate system without Non-Maximum Suppression (NMS).

\subsection{Main Results}
\subsubsection{Results on KITTI} 
In Table \ref{table:kitti car}, we first compare our MonoPixel with state-of-the-art methods on the \textit{Car} class.
We divide state-of-the-art methods into two groups that use a single RGB image and extra input data such as dense depth map or image sequences. 
There is a clear tendency that the depth map approaches outperform RGB image approaches, while the depth map approaches have a significantly slow runtime. 
For example, DORN \cite{Fu2018}, which is a widely adopted depth estimation network for depth map approaches \cite{Weng2019, Ma, Ma2020, Ding2020a}, takes 400$ms$ to predict the dense depth map from a $1280 \times 384$ image.
Depth map approaches suffer from slow speed; thus, the 3D detection network requires additional computational time for the depth estimation network.

Our model achieves state-of-the-art performance on both $AP_{3D}$ and $AP_{BEV}$ metrics while maintaining a fast runtime due to the simple yet effective network design.
Note that SMOKE \cite{Liu}, MonoPair \cite{Chen2020b}, RTM3D \cite{Li2020b}, and MonoDLE \cite{ma2021delving} have single-stage anchor-free architecture following CenterNet \cite{Zhou2019}, but our method outperforms others by a large margin thanks to the pixel-wise depth supervision from LiDAR.
Remarkably, our model also outperforms depth map approaches that use the standalone depth estimation network \cite{Fu2018}, showing the benefit of training depth considering the 3D detection task.
The proposed network runs an order of magnitude faster than depth map approaches since ours does not explicitly predict the dense depth map during the testing phase.
The proposed method also outperforms methods \cite{wang2021depth, reading2021categorical} that use extra data such as LiDAR points or dense depth map during training stage while being much faster.
Unlike CaDDN \cite{reading2021categorical} trained with pixel-wise depth supervision requires to predict dense depth during inference, ours does not predict dense depth during inference which makes the network run 15 times faster.

%%%%%%%%%%%%%%%%%%%%%%%%%%%%%%%%%%%%%%%%%%%%%%%%%%%%%%%%%%%%%%%%%%%%%%%%%%%%%%%%%%%%%%%%%
\setcounter{figure}{5}
\begin{figure*}[!t]
\begin{center}
\includegraphics[width=0.95\textwidth]{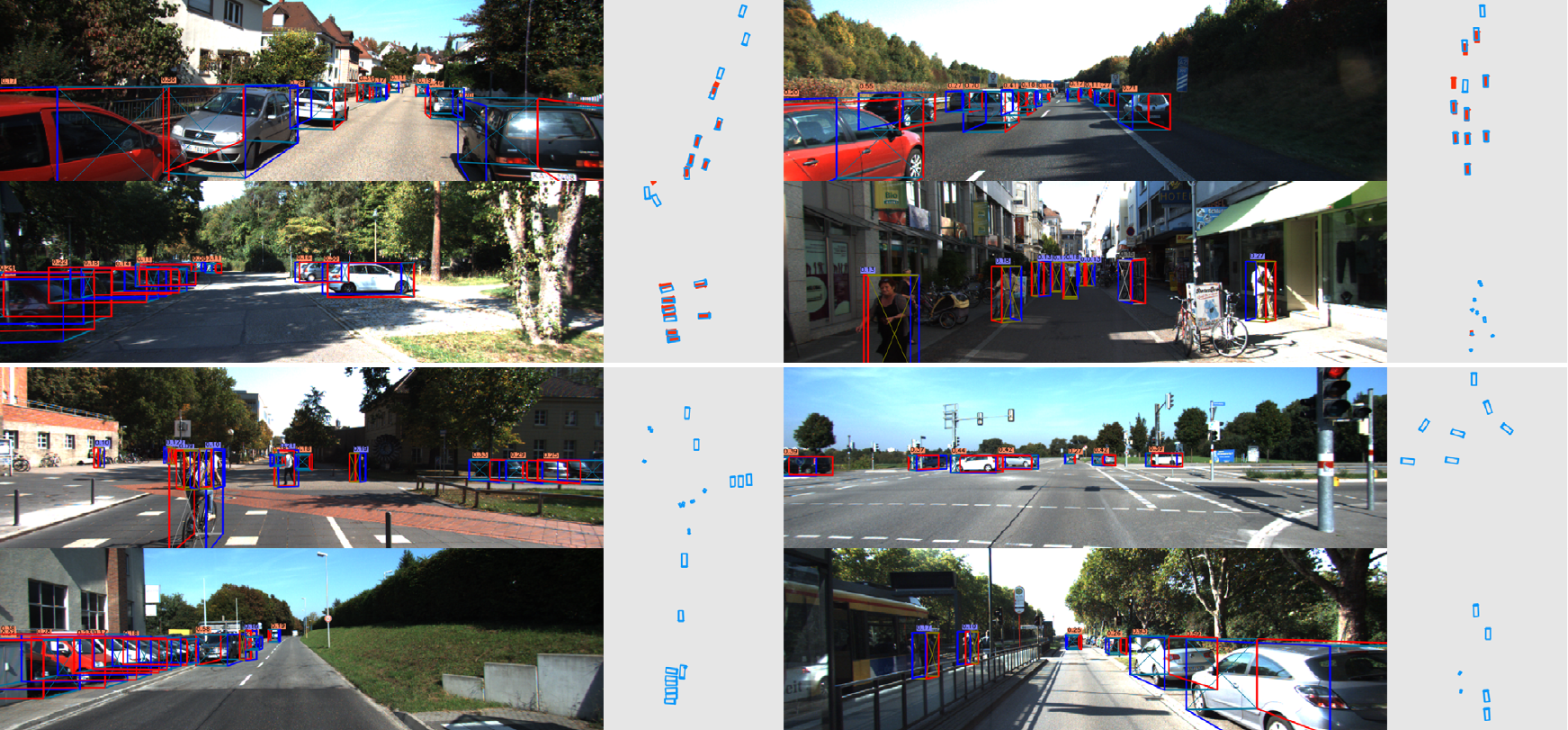} 
\end{center}
\caption{Qualitative results on KITTI (first and second row: \textit{val split}, third and fourth row: \textit{test split}).
The red and blue boxes on the bird's eye view denote ground truths and prediction results.
The maximum range of bird's eye view is 64$m$ and best viewed in color with zoom in.}
\label{fig:kitti figs}
\end{figure*}
%%%%%%%%%%%%%%%%%%%%%%%%%%%%%%%%%%%%%%%%%%%%%%%%%%%%%%%%%%%%%%%%%%%%%%%%%%%%%%%%%%%%%%%%%

%%%%%%%%%%%%%%%%%%%%%%%%%%%%%%%%%%%%%%%%%%%%%%%%%%%%%%%%%%%%%%%%%%%%%%%%%%%%%%%%%%%%%%%%%
\setlength{\tabcolsep}{0.4em} % for the horizontal padding
\setcounter{table}{4}
\begin{table}[!t]
    \caption{$AP_{40}$ Performance on KITTI Test Set for Pedestrian and Cyclist Classes at 0.5 IoU.
    $^*$ Denotes Methods using Depth Map as Network Input.}
    \begin{center}
    \resizebox{1.0\columnwidth}{!}{
    \begin{tabular}{c||ccc|ccc}
        \hline
        \multirow{2}*{Method} & \multicolumn{3}{c|}{Ped., $AP_{3D}$ [\%]} & \multicolumn{3}{c}{Cyc., $AP_{3D}$ [\%]} \\
        & Easy & Mod. & Hard & Easy & Mod. & Hard\\
        \hline
        D4LCN$^*$ \cite{Ding2020a} & 4.55 & 3.42 & 2.83 & 2.45 & 1.67 & 1.36 \\
        Demystifying$^*$ \cite{Simonelli2020a} & 3.00 & 1.81 & 1.59 & 7.79 & 4.32 & 3.98 \\
        \hline
        M3D-RPN \cite{Brazil2019} & 4.92 & 3.48 & 2.94 & 0.94 & 0.65 & 0.47 \\
        MonoPSR \cite{ku2019monocular} & 6.12 & 4.00 & 3.30 & \textbf{8.37} & \underline{4.74} & \underline{3.68} \\
        MoVi-3D \cite{Simonelli2020} & 8.99 & 5.44 & 4.57& 1.08 & 0.63 & 0.70 \\
        MonoDLE \cite{ma2021delving} & 9.64 & 6.55 & 5.44 & 4.59 & 2.66 & 2.45 \\
        MonoPair \cite{Chen2020b} & 10.02 & 6.68 & 5.53 & 3.79 & 2.12& 1.83 \\
        CaDDN \cite{reading2021categorical} & 12.87 & 8.14 & 6.76 & 7.00 & 3.41 & 3.30 \\
        
        \hline
        MonoPixel (Ours) & \textbf{14.01} &	\textbf{9.40} &	\textbf{7.90} &\underline{8.20} & \textbf{4.93} & \textbf{4.02} \\
        \hline
    \end{tabular}}
    \end{center}
    \label{table:kitti ped cyc}
\end{table}
%%%%%%%%%%%%%%%%%%%%%%%%%%%%%%%%%%%%%%%%%%%%%%%%%%%%%%%%%%%%%%%%%%%%%%%%%%%%%%%%%%%%%%%%%

Moreover, the performance of \textit{Pedestrian} and \textit{Cyclist} classes in Table \ref{table:kitti ped cyc} highlights the benefit of our object-centric auxiliary depth loss.
Note that only the \textit{Car} class is evaluated and \textit{Pedestrian} and \textit{Cyclist} classes are often excluded in other methods due to the imbalanced class distribution (\textit{i.e.}, the KITTI training split consists of 14357, 2207, and 734 labels of Car, Pedestrian, and Cyclist classes, respectively.
On the \textit{Pedestrian} class, RGB image approaches surpass depth map approaches in contrast to the \textit{Car} class.
We argue that the number of depth supervision (LiDAR points) on vulnerable road users is too small to train the depth estimation network, and depth map approaches fail to be trained to estimate the accurate position on these classes.
On \textit{Cyclist}, we outperform the previous leading method, MonoPSR \cite{ku2019monocular}, which leverages depth supervisions from LiDAR points as 3D shape reconstruction loss.
However, our method runs 5 times faster than \cite{ku2019monocular} thanks to the efficient single-stage architecture, as opposed to \cite{ku2019monocular} having multiple region proposal and refinement networks.

As the qualitative results shown in Fig. \ref{fig:kitti figs}, our network predicts tight bounding boxes on multi-class objects in challenging scenarios.
Objects that are distant and occluded are difficult to accurately be detected by a monocular-based 3D object detector.
However, our network not only improves the overall localization performance by the proposed object-centric depth prediction loss, but further adjusts the confidence score of difficult objects using the confidence of estimated depth.

\subsubsection{Results on nuScenes} 
In Table \ref{table:ns results}, we compare our MonoPixel with CenterNet \cite{Zhou2019} and FCOS3D \cite{wang2021fcos3d} on the nuScenes validation set using $AP$ at various distance thresholds and $TP$ metrics at a 2$m$ threshold.
The results of CenterNet and FCOS3D are from the official code\renewcommand{\thefootnote}{4}\footnotemark\renewcommand{\thefootnote}{5}\footnotemark and model provided by the authors.
Our method gains significant $AP$ improvement compared to CenterNet across all classes and thresholds, especially effective on strict thresholds (\textit{i.e.}, 0.5$m$ or 1.0$m$).
On three major classes, Car, Pedestrian, and Bicycle, our method improves the CenterNet by 5.15\%, 2.24\%, and 7.37\% $AP$, respectively, at a 1.0$m$ distance threshold.
The performance gain on less frequently appearing classes are larger, which demonstrates the effectiveness of pixel-wise supervision when the instance-wise supervision is insufficient.
Our method also outperforms FCOS3D except 4.0$m$ threshold with much faster inference speed.
Note that frame per second (FPS) of FCOS3D is 1.7, while ours is 4.7.
Particularly, we found that the reduced translation error is the major factor of AP improvement.
Since the positive matching criteria of nuScenes is related to the center distance between the ground truth and prediction, our method can have a higher recall by improved localization performance.
Interestingly, the orientation error is also significantly improved by the proposed method, assuming that training pixel-wise depth can benefit the network to estimate the object's orientation.
We also visualize multi-class prediction results of six surrounding cameras for 360 degree detection on nuScenes as shown in Fig. \ref{fig:ns figs}.

%%%%%%%%%%%%%%%%%%%%%%%%%%%%%%%%%%%%%%%%%%%%%%%%%%%%%%%%%%%%%%%%%%%%%%%%%%%%%%%%%%%%%%%%%
\renewcommand{\thefootnote}{4}
\footnotetext[4]{: \url{https://github.com/xingyizhou/CenterTrack}}
\renewcommand{\thefootnote}{5}
\footnotetext[5]{: \url{https://github.com/open-mmlab/mmdetection3d}}
%%%%%%%%%%%%%%%%%%%%%%%%%%%%%%%%%%%%%%%%%%%%%%%%%%%%%%%%%%%%%%%%%%%%%%%%%%%%%%%%%%%%%%%%%

%%%%%%%%%%%%%%%%%%%%%%%%%%%%%%%%%%%%%%%%%%%%%%%%%%%%%%%%%%%%%%%%%%%%%%%%%%%%%%%%%%%%%%%%%
\setcounter{figure}{6}
\begin{figure*}[!t]
\begin{center}
\includegraphics[width=0.95\textwidth]{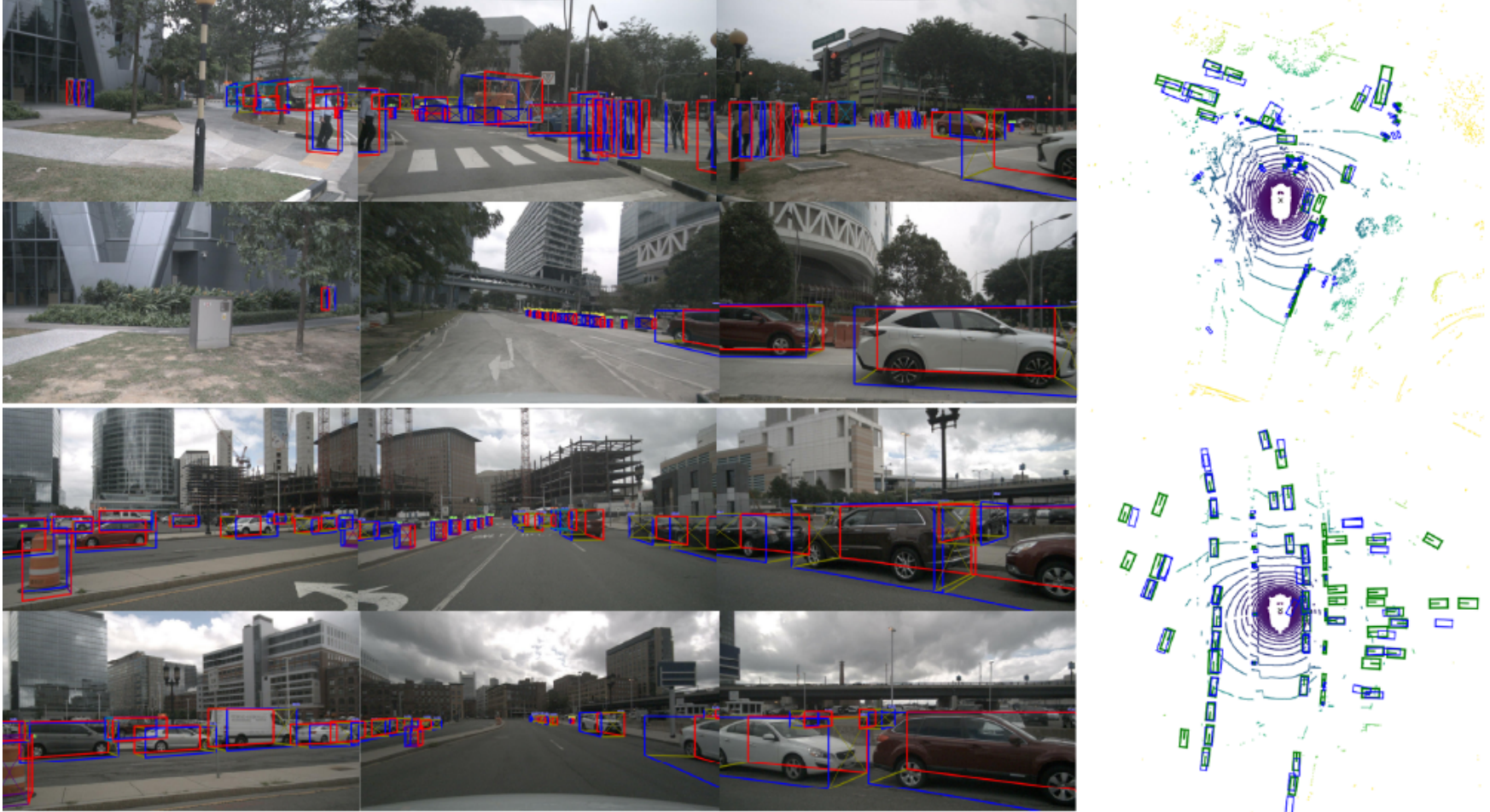} 
\end{center}
\caption{Qualitative results of six surrounding camera images on nuScenes \textit{val set}.
The green and blue boxes on the bird's eye view denote ground truths and prediction results. 
Note that three rear camera images are flipped for the better visualization and figures are best viewed in color with zoom in.}
\label{fig:ns figs}
\end{figure*}
%%%%%%%%%%%%%%%%%%%%%%%%%%%%%%%%%%%%%%%%%%%%%%%%%%%%%%%%%%%%%%%%%%%%%%%%%%%%%%%%%%%%%%%%%

%%%%%%%%%%%%%%%%%%%%%%%%%%%%%%%%%%%%%%%%%%%%%%%%%%%%%%%%%%%%%%%%%%%%%%%%%%%%%%%%%%%%%%%%%
\setlength{\tabcolsep}{0.23em}
\setcounter{table}{5}
\begin{table}[!t]
    \caption{Performance on nuScenes Val Set.
    All Denotes the Average Performance over 10 Classes.}
    \begin{center}
    \resizebox{1.0\columnwidth}{!}{
    \begin{tabular}{c|c||cccc|ccc}
        \hline
        \multirow{2}{*}{Class} & \multirow{2}{*}{Method} & \multicolumn{4}{c|}{$AP$$\uparrow$ [$\%$]} & ATE$\downarrow$ & ASE$\downarrow$ & AOE$\downarrow$ \\
        & & 0.5$m$ & 1.0$m$ & 2.0$m$ & 4.0$m$ & [$m$] & [1-IoU] & [rad]\\
        \hline
        \multirow{3}{*}{Car} 
        & CenterNet & 17.28 & 40.22 & 60.69 & 72.89 & 0.498 & 0.146 & 0.089 \\
        & FCOS3D    & 12.86 & 36.38 & 63.54 & 78.91 & 0.569 & 0.149 & 0.112 \\
        & MonoPixel & 19.58 & 45.37 & 66.75 & 77.82 & 0.488 & 0.163 & 0.065 \\
        \hline
        \multirow{3}{*}{Ped.} 
        & CenterNet &  8.43 & 27.48 & 49.24 & 64.55 & 0.614 & 0.282 & 0.853 \\
        & FCOS3D    &  9.64 & 30.06 & 54.31 & 70.47 & 0.667 & 0.286 & 0.714 \\
        & MonoPixel &  9.86 & 29.72 & 51.48 & 66.31 & 0.586 & 0.299 & 0.491 \\
        \hline
        \multirow{3}{*}{Bic.} 
        & CenterNet & 2.80 & 14.72 & 29.01 & 38.40 & 0.667 & 0.276 & 0.972 \\
        & FCOS3D    & 5.77 & 22.23 & 41.11 & 51.59 & 0.653 & 0.264 & 0.673 \\
        & MonoPixel & 6.03 & 22.09 & 42.59 & 48.59 & 0.630 & 0.274 & 0.714 \\
        \hline
        \multirow{3}{*}{All} 
        & CenterNet & 8.02 & 21.72 & 38.33 & 52.06 & 0.703 & 0.258 & 0.600 \\
        & FCOS3D    & 7.91 & 22.45 & 41.31 & 57.10 & 0.738 & 0.261 & 0.487 \\
        & MonoPixel & 9.35 & 24.20 & 43.46 & 55.67 & 0.692 & 0.271 & 0.425 \\
        \hline
    \end{tabular}}
    \end{center}
    \label{table:ns results}
\end{table}
%%%%%%%%%%%%%%%%%%%%%%%%%%%%%%%%%%%%%%%%%%%%%%%%%%%%%%%%%%%%%%%%%%%%%%%%%%%%%%%%%%%%%%%%%

\subsection{Ablation Study} \label{Ablation Study}

For ablation studies, we report the $AP_{40}$ of the \textit{Car} class on the KITTI \textit{val split} at 0.7 IoU  \textit{moderate} difficulty, and models are trained on the KITTI \textit{train split} with the hyperparameters described in Sec. \ref{Implementation Details} unless mentioned otherwise.

%%%%%%%%%%%%%%%%%%%%%%%%%%%%%%%%%%%%%%%%%%%%%%%%%%%%%%%%%%%%%%%%%%%%%%%%%%%%%%%%%%%%%%%%%
\setcounter{table}{6}
\setlength{\tabcolsep}{0.35em}
\begin{table*}[!t]
    \caption{Object-Centric / Raw Depth Performance Comparison of Car Class by Object-Centric Auxiliary Depth Loss}
    \begin{center}
    \resizebox{0.85\textwidth}{!}
    {
    \begin{tabular}{c|c||c|cccccc}
        \hline
         & \multirow{2}{*}{Metric} & \multirow{2}{*}{DORN \cite{Fu2018}} & \multicolumn{6}{c}{Foreground Mask Weight $\lambda$} \\
        & & & 0 & 0.3 & 0.5 & 0.7 & 0.8 & 1.0\\
        \hline
        \multirowcell{3}{Lower\\is\\better} 
        &Abs Rel &0.170/\textbf{0.109}& 0.205/0.127 &0.152/0.124 &0.130/0.128 &0.118/0.134 &0.114/0.138 &\textbf{0.107}/0.174\\
        &Sq Rel  &1.125/\textbf{0.677}& 2.154/0.890 &0.963/0.799 &0.631/0.833 &0.497/0.923 &0.449/0.971 &\textbf{0.404}/1.449\\
        &RMSE    &3.256/\textbf{3.635}& 4.750/3.968 &3.192/3.951 &2.634/4.097 &2.287/4.307 &2.154/4.426 &\textbf{1.985}/5.963\\
        \hline
        \multirowcell{5}{Higher\\is\\better} 
        &$\delta<$1.10  &0.714/\textbf{0.716} &0.661/0.666 &0.727/0.667 &0.748/0.661 &0.764/0.652 &0.773/0.646 &\textbf{0.793}/0.560\\
        &$\delta<$1.25  &0.868/\textbf{0.892} &0.828/0.853 &0.866/0.849 &0.887/0.842 &0.902/0.832 &0.908/0.828 &\textbf{0.916}/0.752\\
        &$\delta<1.25^2$&0.925/\textbf{0.961} &0.904/0.949 &0.916/0.944 &0.929/0.938 &0.939/0.930 &0.939/0.926 &\textbf{0.940}/0.869\\
        \cline{2-9}
        & $AP_{3D}$ & - &18.93 &20.41 &20.63 &\textbf{20.99} &20.81 &20.72\\
        & $AP_{BEV}$& - &25.56 &26.29 &26.67 &27.10 &\textbf{27.31} &26.13\\
        \hline
    \end{tabular}
    }
    \end{center}
    \label{table:object-centric ablation}
\end{table*}
%%%%%%%%%%%%%%%%%%%%%%%%%%%%%%%%%%%%%%%%%%%%%%%%%%%%%%%%%%%%%%%%%%%%%%%%%%%%%%%%%%%%%%%%%

\subsubsection{Object-Centric Auxiliary Depth Loss}  
To show the effectiveness of our object-centric auxiliary depth loss, we evaluate the \textit{object-centric} and raw depth estimation performance in addition to $AP$ performance in Table \ref{table:object-centric ablation}.
Object-centric depth estimation performance considers the depth error that occurred in the foreground pixels (Car class), while raw depth estimation performance considers the depth error on every ground truth pixel (all projected LiDAR points).
The results show that the object-centric depth estimation performance is improved by increasing the foreground mask weight for the object-centric depth loss and outperforms DORN \cite{Fu2018}.
The $AP$ performance is also improved by improved object-centric depth estimation performance, and we found $\lambda$=0.7 to work best in our experiments.
This suggests the 3D detection network may benefit from depth information, not only from the depths of foreground objects, but also the background.

\subsubsection{Object Depth Decomposition}
We evaluate the performance of our method with and without the object depth decomposition strategy in Table \ref{table:performance by ODD}.
Aux. Dep. and ODD denote the auxiliary depth loss and object depth decomposition strategy, respectively.
The model without the object depth decomposition strategy trains the instance-wise depth loss (Eq. \ref{eq:instance-wise depth loss}) with the object's center depth and trains the pixel-wise depth loss (Eq. \ref{eq:pixel-wise depth loss}) with the object's surface depth.
We found that the object depth decomposition improves performance only when the network is trained by pixel-wise supervision, showing that our decomposition strategy allows the network to leverage the depth value from LiDAR points more effectively.
The improvement is more significant in \textit{Pedestrian} and \textit{Cyclist} classes, which have much fewer training samples compared to the \textit{Car} class.
The results verify that the depth decomposition strategy enables the network to make the best use of per-pixel depth supervisions, especially when the instance-wise depth supervision is insufficient.

%%%%%%%%%%%%%%%%%%%%%%%%%%%%%%%%%%%%%%%%%%%%%%%%%%%%%%%%%%%%%%%%%%%%%%%%%%%%%%%%%%%%%%%%%
\setcounter{table}{7}
\setlength{\tabcolsep}{0.4em}
\begin{table*}[!t]
    \caption{Effects of Object Depth Decomposition Strategy}
    \begin{center}
    \resizebox{0.62\textwidth}{!}
    {
    \begin{tabular}{cc||ccc|ccc|ccc}
        \hline
        \multirowcell{2}{Aux.\\Dep.} & \multirow{2}*{ODD} & \multicolumn{3}{c|}{Car, $AP_{3D}$ [\%]} & \multicolumn{3}{c|}{Ped., $AP_{3D}$ [\%]} & \multicolumn{3}{c}{Cyc., $AP_{3D}$ [\%]} \\
        & & Easy & Mod. & Hard & Easy & Mod. & Hard & Easy & Mod. & Hard\\
        \hline
        &                    & 22.82 & 18.06 & 16.87 & 10.47 & 8.53  & 6.68 & 12.33 & 6.54 & 6.12 \\
        & \checkmark         & 22.30 & 17.88 & 16.44 & 10.94 & 8.04  & 6.73 & 12.47 & 6.65 & 6.00 \\
        \checkmark &         & 25.50 & 20.73 & 18.48 & 11.88 & 9.18  & 7.75 & 12.01 & 6.96 & 6.34 \\
        \checkmark&\checkmark& \textbf{25.69} & \textbf{21.00} & \textbf{18.69} & \textbf{12.70} & \textbf{10.20} & \textbf{8.23} & \textbf{14.07} & \textbf{7.21} & \textbf{6.99} \\
        \hline
    \end{tabular}}
    \end{center}
    \label{table:performance by ODD}
\end{table*}
%%%%%%%%%%%%%%%%%%%%%%%%%%%%%%%%%%%%%%%%%%%%%%%%%%%%%%%%%%%%%%%%%%%%%%%%%%%%%%%%%%%%%%%%%

%%%%%%%%%%%%%%%%%%%%%%%%%%%%%%%%%%%%%%%%%%%%%%%%%%%%%%%%%%%%%%%%%%%%%%%%%%%%%%%%%%%%%%%%%
\setcounter{table}{8}
\setlength{\tabcolsep}{0.35em}
\begin{table}[!t]
    \caption{Effects of Depth Uncertainty}
    \begin{center}
    \resizebox{1.0\columnwidth}{!}
    {
    \begin{tabular}{cc||ccc|ccc|c}
        \hline
        \multirowcell{2}{Aux.\\Dep.} & \multirowcell{2}{Dep.\\Unc.} & \multicolumn{3}{c|}{Car, $AP_{3D}$ [\%]} & \multicolumn{3}{c|}{Car, $AP_{BEV}$ [\%]} & Time\\
        & &  Easy & Mod. & Hard & Easy & Mod. & Hard & [$ms$] \\
        \hline
        &                    & 22.05&	18.61&	16.53&	30.57&	25.36&	23.15 & \textbf{40.4}\\
        &\checkmark          & 21.22&	17.72&	16.26&	28.57&	24.18&	22.64 & 42.3 \\
        \checkmark      &   & 25.35&	20.68&	18.51&	32.59&	26.35&	23.92 & 40.6 \\
        \checkmark&\checkmark& \textbf{25.47}&	\textbf{20.90}&	\textbf{18.81}&	\textbf{33.60}&	\textbf{27.15}&	\textbf{25.56} & 42.4 \\
        \hline
    \end{tabular}}
    \end{center}
    \label{table:performance by component}
\end{table}
%%%%%%%%%%%%%%%%%%%%%%%%%%%%%%%%%%%%%%%%%%%%%%%%%%%%%%%%%%%%%%%%%%%%%%%%%%%%%%%%%%%%%%%%%

\subsubsection{Depth Uncertainty Guided 3D Object Confidence}  
We conduct ablation experiments by adding auxiliary depth loss and depth uncertainty components in Table \ref{table:performance by component}.
Our depth uncertainty significantly improves the detection performance by adding only 2$ms$ of computational overhead.
We found that using depth uncertainty without auxiliary depth loss (train depth-related sub-networks with instance-wise loss only) does not improve the performance, but it is highly effective when trained pixel-wise loss. 
We suppose the instance-wise loss does not provide enough supervision to generalize the depth uncertainty sub-network.
Moreover, the performance gain of using auxiliary depth loss is higher on $AP_{BEV}$ than $AP_{3D}$ since our method focuses on improving the depth in a radial direction, not in a vertical direction.

%%%%%%%%%%%%%%%%%%%%%%%%%%%%%%%%%%%%%%%%%%%%%%%%%%%%%%%%%%%%%%%%%%%%%%%%%%%%%%%%%%%%%%%%%
\setcounter{figure}{7}
\begin{figure}[t]
\begin{center}
\includegraphics[width=0.95\columnwidth]{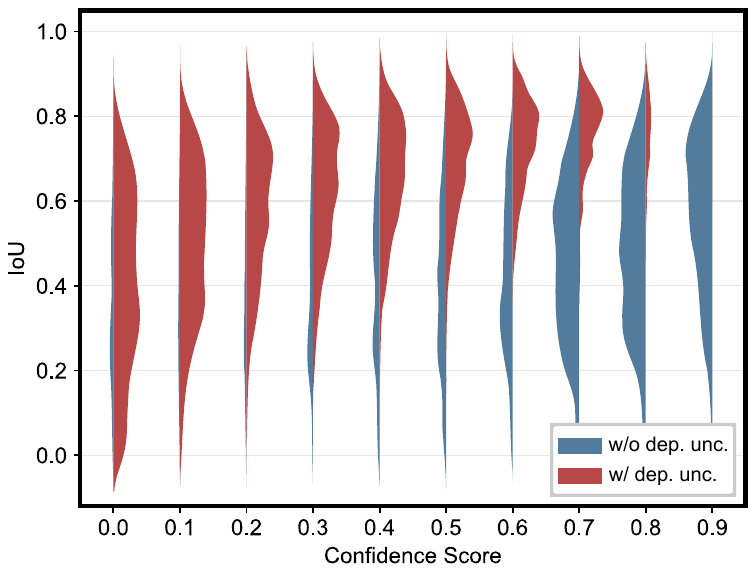}
\end{center}
\caption{Distribution of 3D IoU over confidence scores between prediction and ground truth.
2$m$ center-distance threshold is used for matching.}
\label{fig:IoU over conf}
\end{figure}
%%%%%%%%%%%%%%%%%%%%%%%%%%%%%%%%%%%%%%%%%%%%%%%%%%%%%%%%%%%%%%%%%%%%%%%%%%%%%%%%%%%%%%%%%

We additionally compare the distribution of 3D IoU by the confidence scores in Fig. \ref{fig:IoU over conf} to show the effect of the depth uncertainty guided 3D object confidence.
The results without depth uncertainty denotes the 2D confidence score trained by general cross-entropy loss defined in the 2D image plane in Eq. \ref{eq:keypoint loss}.
Without considering the localization uncertainty in the depth direction, image-based methods are prone to be overconfident if objects provide rich visual cues (\textit{e.g.}, clearly visible and easily distinguishable objects from the background).
However, predictions with high 2D confidence often fail to be accurately localized in 3D space; thus, even high confident prediction can have a low 3D IoU with the ground truth.
On the other hand, with our depth uncertainty guided 3D confidence, the 3D confidence is reduced when the prediction has high localization uncertainty, although the prediction has high 2D confidence.
As a result, the proposed 3D confidence with depth uncertainty becomes high only when the prediction has high localization accuracy.

%-------------------------------------------------------------------------
\section{Conclusion}
In this paper, we introduced a novel monocular image-based 3D object detector with object-centric auxiliary depth loss to improve the depth estimation performance for 3D detection.
Our method leverages the pixel-wise depth supervision from raw LiDAR points which can be obtained without human annotation cost.
The proposed loss can provide large-scale depth information to the 3D detection network without requiring additional computational cost during inference time.
Our method, trained with object-centric auxiliary depth loss, has significantly improved depth estimation performance and even outperforms depth map approaches that use the standalone depth estimation network. 
On the KITTI and nuScenes benchmark, our MonoPixel achieves competitive performance with state-of-the-art monocular-based 3D detection methods while maintaining real-time inference speed.
% We believe that the proposed simple, yet effective auxiliary depth loss for the 3D object detector can be generally adapted to other baselines.
%---------------------------- ---------------------------------------------

\bibliographystyle{IEEEtran}
\bibliography{IEEEfull}

\begin{IEEEbiography}[{\includegraphics[width=1in,height=1.25in,clip,keepaspectratio]{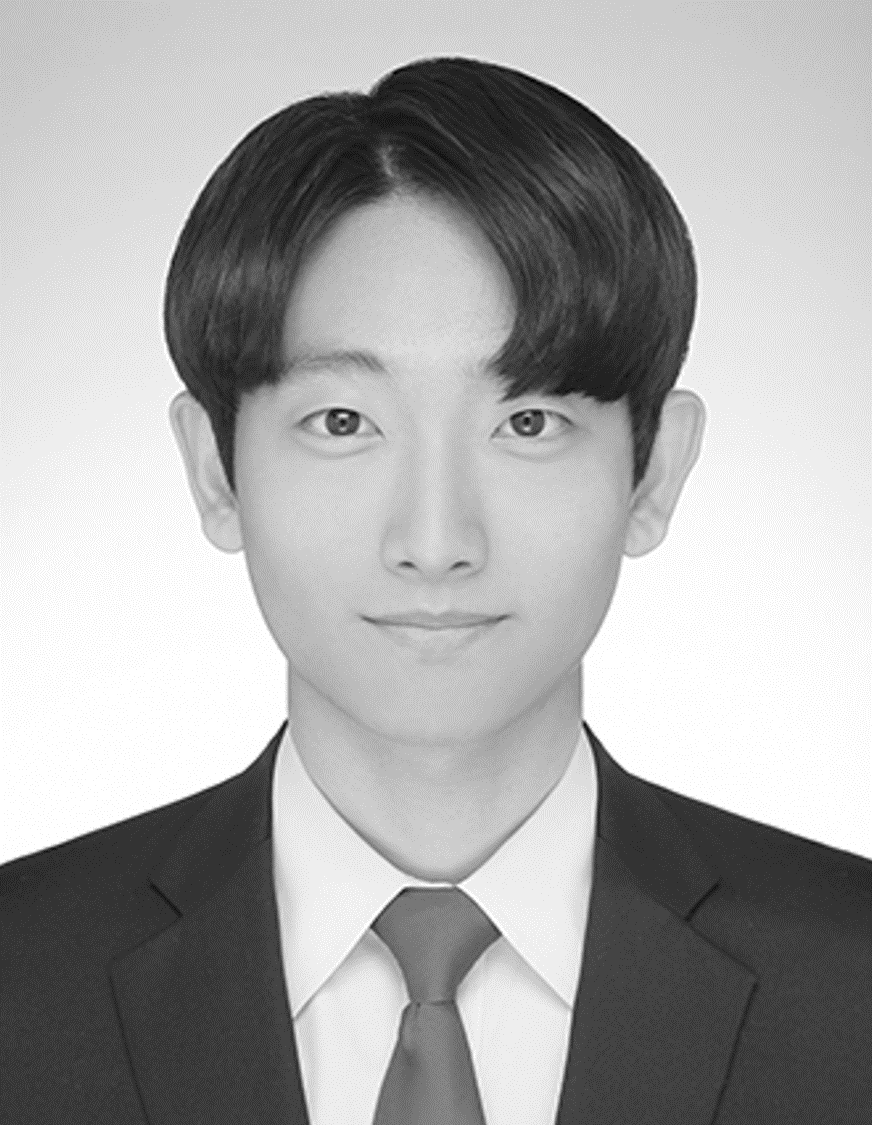}}]{Youngseok Kim}
received a B.S. degree in Mechatronics Engineering from the Korea University of Technology and Education (KoreaTech), Cheonan, South Korea, in 2017, and an M.S. degree from the Graduate School of Mobility, Korea Advanced Institute of Science and Technology (KAIST), Daejeon, South Korea, in 2019. He is currently pursuing a Ph.D. degree in the Graduate School of Mobility, KAIST. His research interests include 3D computer vision for autonomous driving.
\end{IEEEbiography}

\begin{IEEEbiography}[{\includegraphics[width=1in,height=1.25in,clip,keepaspectratio]{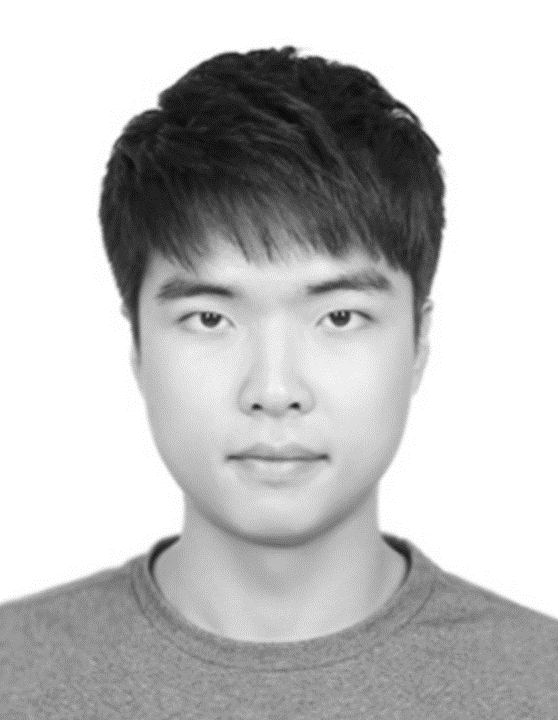}}]{Sanmin Kim}
received a B.S degree in Mechanical Engineering and M.S degree in the Graduate School of Mobility from the Korea Advanced Institute of Science and Technology (KAIST), Daejeon, South Korea, in 2018 and 2020, respectively. He is currently pursuing a Ph.D. degree in the Graduate School of Mobility, KAIST. His research interests include deep learning for perception and trajectory prediction in autonomous driving vehicles.
\end{IEEEbiography}

\begin{IEEEbiography}[{\includegraphics[width=1in,height=1.25in,clip,keepaspectratio]{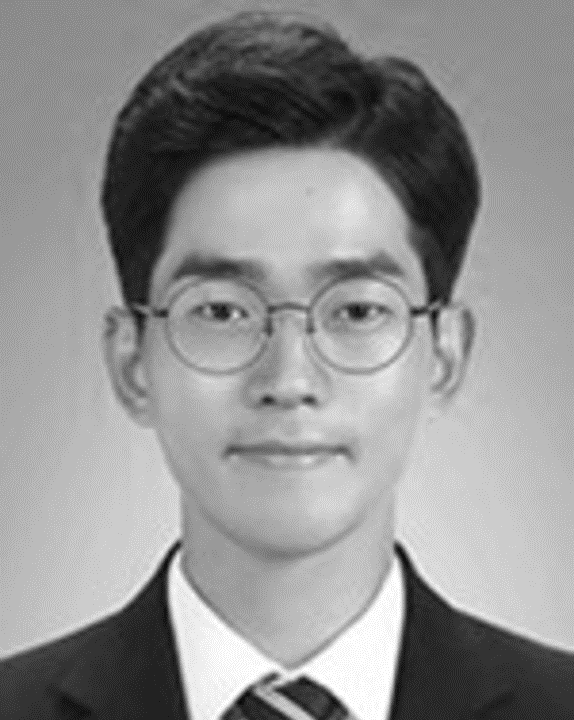}}]{Sangmin Sim}
received a B.S. degree in Mechanical Engineering from the Hongik University, Seoul, South Korea, in 2020, and an M.S. degree from the Graduate School of Mobility, Korea Advanced Institute of Science and Technology (KAIST), Daejeon, South Korea, in 2022. He is currently working as a researcher for Hyundai Motor Company. His research interests include deep learning for perception and 3D computer vision.
\end{IEEEbiography}

\begin{IEEEbiography}[{\includegraphics[width=1in,height=1.25in,clip,keepaspectratio]{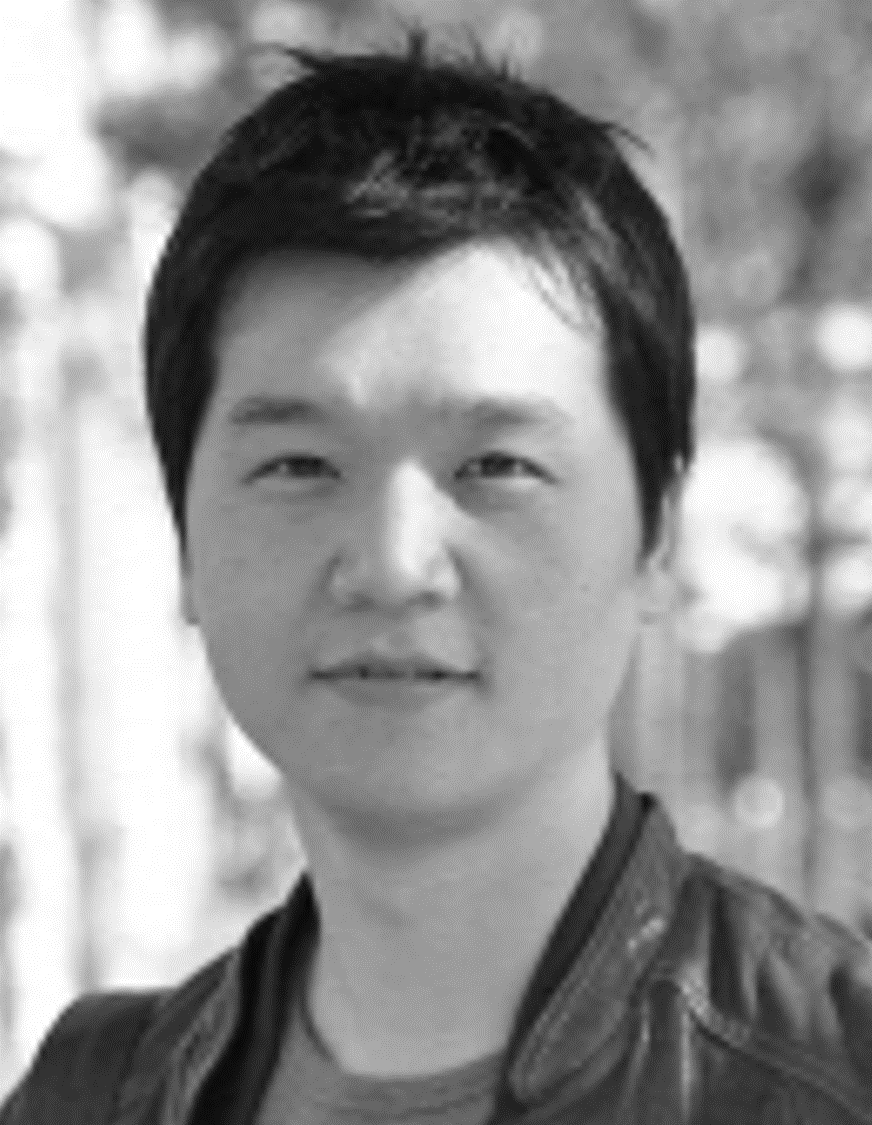}}]{Dongsuk Kum}
received a Ph.D. degree in mechanical engineering from the University of Michigan, Ann Arbor, MI, USA, in 2010. He was a Visiting Research Scientist with the General Motors Research and Development Propulsion Systems Research Laboratory, Warren, MI, USA, where he focused on advanced propulsion system technologies, including hybrid electric vehicles, flywheel hybrid, and waste heat recovery systems. He is currently an Associate Professor with the Graduate School of Mobility, Korea Advanced Institute of Science and Technology, where he is also the Director of the Vehicular Systems Design and Control Laboratory. His research centers on the modeling, control, and design of advanced vehicular systems with particular interests in hybrid electric vehicles and autonomous vehicles.
\end{IEEEbiography}

\begin{IEEEbiography}[{\includegraphics[width=1in,height=1.25in,clip,keepaspectratio]{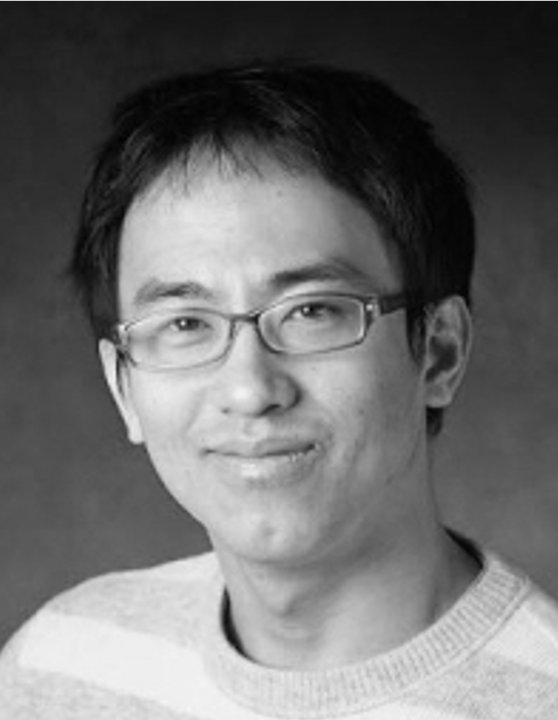}}]{Junwon Choi}
received B.S. and M.S. degrees from the Electrical Engineering Department, Seoul National University, and a Ph.D. degree in electrical computer engineering from the University of Illinois at Urbana–Champaign. In 2010, he joined Qualcomm, San Diego, USA, where he participated in research on advanced signal processing technology for next-generation wireless systems. Since 2013, he has been with the Electrical Engineering Department, Hanyang University, as a Faculty Member. His research area includes signal processing, machine learning, intelligent vehicles, and wireless communications.
\end{IEEEbiography}

\vfill

\end{document}